\documentclass{article}

\usepackage[preprint]{neurips_2025}

\usepackage[utf8]{inputenc} 
\usepackage[T1]{fontenc}    
\usepackage{hyperref}       
\usepackage{url}            
\usepackage{booktabs}       
\usepackage{amsfonts}       
\usepackage{nicefrac}       
\usepackage{microtype}      
\usepackage{xcolor}         
\usepackage{subcaption}   
\usepackage{svg}
\usepackage{wrapfig}

\usepackage{amsmath}
\usepackage{amssymb}
\usepackage{acronym}

\usepackage{graphicx}
\usepackage{multirow}
\usepackage{algorithm}
\usepackage{algorithmic}

\usepackage{bm}
\usepackage{subcaption}

\title{Automatic mixed precision for optimizing gained time with constrained loss mean-squared-error based on model partition to sequential sub-graphs}

\author{ 
{Shmulik Markovich-Golan}
	\And
{Daniel Ohayon}
	\And
{Itay Niv} 
    \And
{Yair Hanani}     \And
\\
Intel Corporation/Habana\\ 
\{shmulik.markovich-golan, daniel1.ohayon, itay.niv, yair.hanani\}@intel.com}

\date{Intel Corporation/Habana\\ 
\{shmulik.markovich-golan/daniel1.ohaion/itay.niv/yair.hanani\}@intel.com}

\begin{document}
\newcommand{\vX}{\mathbf{X}}
\newcommand{\vY}{\mathbf{Y}}
\newcommand{\vXl}{\mathbf{X}_{\ell}}
\newcommand{\vXalt}{\mathbf{X}_{0,\ell'}}
\newcommand{\vXblt}{\mathbf{X}_{1, \ell'}}
\newcommand{\vXal}{\mathbf{X}_{0,\ell}}
\newcommand{\vXbl}{\mathbf{X}_{1, \ell}}
\newcommand{\vWl}{\mathbf{W}_{\ell}}
\newcommand{\vWlT}{\mathbf{W}_{\ell}^{T}}
\newcommand{\vYl}{\mathbf{Y}_{\ell}}
\newcommand{\vylt}{\mathbf{y}_{\ell'}}
\newcommand{\vYlt}{\mathbf{Y}_{\ell'}}
\newcommand{\vbl}{\mathbf{b}_{\ell}}
\newcommand{\vblT}{\mathbf{b}_{\ell}^{T}}
\newcommand{\Cl}{C_{\ell}}
\newcommand{\Kl}{K_{\ell}}
\newcommand{\vone}{\mathbf{1}}
\newcommand{\ven}{\mathbf{e}_{n}}
\newcommand{\venT}{\mathbf{e}_{n}^{T}}
\newcommand{\Clt}{C_{\ell'}}
\newcommand{\Llin}{\mathcal{L}_{\textrm{lin}}}
\newcommand{\LBGEMM}{\mathcal{L}_{\textrm{BGEMM}}}
\newcommand{\hvXl}{\hat{\mathbf{X}}_{\ell}}
\newcommand{\hvXlta}{\hat{\mathbf{X}}_{\ell',0}}
\newcommand{\hvXltab}{\hat{\mathbf{X}}_{\ell',1}}
\newcommand{\hvWl}{\hat{\mathbf{W}}_{\ell}}
\newcommand{\tvXl}{\tilde{\mathbf{X}}_{\ell}}
\newcommand{\tvXlta}{\tilde{\mathbf{X}}_{\ell',0}}
\newcommand{\tvXltab}{\tilde{\mathbf{X}}_{\ell',1}}
\newcommand{\tvWl}{\tilde{\mathbf{W}}_{\ell}}
\newcommand{\Ilinnl}{I_{\textrm{lin},\ell}^{n}}
\newcommand{\IBGEMMnlt}{I_{\textrm{BGEMM},\ell'}^{n}}
\newcommand{\hg}{\hat{g}}
\newcommand{\tg}{\tilde{g}}
\newcommand{\gj}{g^{r}}
\newcommand{\hgj}{\hat{g}^{r}}
\newcommand{\tgj}{\tilde{g}^{r}}
\newcommand{\vwl}{\mathbf{w}_{\ell}}
\newcommand{\hvwl}{\hat{\mathbf{w}}_{\ell}}
\newcommand{\tvwl}{\tilde{\mathbf{w}}_{\ell}}
\newcommand{\hvxl}{\hat{\mathbf{x}}_{\ell}}
\newcommand{\vxl}{\mathbf{x}_{\ell}}
\newcommand{\tvxl}{\tilde{\mathbf{x}}_{\ell}}
\newcommand{\hvXalt}{\hat{\mathbf{X}}_{0,\ell'}}
\newcommand{\hvXblt}{\hat{\mathbf{X}}_{1, \ell'}}
\newcommand{\tvXalt}{\tilde{\mathbf{X}}_{0,\ell'}}
\newcommand{\tvXblt}{\tilde{\mathbf{X}}_{1, \ell'}}
\newcommand{\hvxalt}{\hat{\mathbf{x}}_{0,\ell'}}
\newcommand{\hvxblt}{\hat{\mathbf{x}}_{1, \ell'}}
\newcommand{\tvxalt}{\tilde{\mathbf{x}}_{0,\ell'}}
\newcommand{\tvxblt}{\tilde{\mathbf{x}}_{1, \ell'}}
\newcommand{\vxalt}{\mathbf{x}_{0,\ell'}}
\newcommand{\vxblt}{\mathbf{x}_{1, \ell'}}
\newcommand{\vect}{\textrm{vec}}
\newcommand{\axli}{\left|x_{\ell,i}\right|}
\newcommand{\awli}{\left|w_{\ell,i}\right|}
\newcommand{\axalti}{\left|x_{0,\ell',i}\right|}
\newcommand{\axblti}{\left|x_{1,\ell',i}\right|}
\newcommand{\txli}{\tilde{x}_{\ell,i}}
\newcommand{\twli}{\tilde{w}_{\ell,i}}
\newcommand{\txalti}{\tilde{x}_{0,\ell',i}}
\newcommand{\txblti}{\tilde{x}_{1,\ell',i}}
\newcommand{\E}{\textrm{E}}
\newcommand{\sigmasq}{\sigma^2}
\newcommand{\dxli}{\dot{x}_{\ell,i}}
\newcommand{\dwli}{\dot{w}_{\ell,i}}
\newcommand{\dxalti}{\dot{x}_{0,\ell',i}}
\newcommand{\dxblti}{\dot{x}_{1,\ell',i}}

\newcommand{\vxj}{\mathbf{x}^{j}}
\newcommand{\vxlj}{\mathbf{x}_{\ell}^{j}}
\newcommand{\dvxlj}{\dot{\mathbf{x}}_{\ell}^{j}}
\newcommand{\dvwlj}{\dot{\mathbf{w}}_{\ell}^{j}}
\newcommand{\dvxaltj}{\dot{\mathbf{x}}_{0,\ell'}^{j}}
\newcommand{\dvxbltj}{\dot{\mathbf{x}}_{1, \ell'}^{j}}
\newcommand{\vxaltj}{\mathbf{x}_{0,\ell'}^{j}}
\newcommand{\vxbltj}{\mathbf{x}_{1, \ell'}^{j}}
\newcommand{\tvxlj}{\tilde{\mathbf{x}}_{\ell}^{j}}
\newcommand{\tvxaltj}{\tilde{\mathbf{x}}_{0,\ell'}^{j}}
\newcommand{\tvxbltj}{\tilde{\mathbf{x}}_{1, \ell'}^{j}}

\newcommand{\tsigmagsq}{\tilde{\sigma}_{g}^2}

\newcommand{\N}{\mathcal{N}}
\newcommand{\slinl}{s_{\textrm{lin},\ell}}
\newcommand{\sBGEMMlt}{s_{\textrm{BGEMM},\ell'}}
\newcommand{\M}{\mathcal{M}}
\newcommand{\plinl}{p_{\textrm{lin},\ell}}
\newcommand{\pBGEMMlt}{p_{\textrm{BGEMM},\ell'}}

\newcommand{\Clsq}{C^2_{\ell}}
\newcommand{\ilf}{i_{\ell,f}}
\newcommand{\dlf}{d_{\ell,f}}
\newcommand{\deltalf}{\delta_{\ell,f}}
\newcommand{\cI}{\mathcal{I}}
\newcommand{\cuI}{\mathcal{\underline{I}}}
\newcommand{\uilf}{\underline{i}_{\ell,f}}
\newcommand{\txtE}{\textrm{E}}
\newcommand{\sigmafsq}{\sigma_f^2}
\newcommand{\mf}{m_{f}}

\newcommand{\axaltisq}{\left|x_{0,\ell',i}\right|^2}
\newcommand{\axbltisq}{\left|x_{1,\ell',i}\right|^2}
\newcommand{\txlisq}{\tilde{x}_{\ell,i}^2}
\newcommand{\twlisq}{\tilde{w}_{\ell,i}^2}
\newcommand{\txaltisq}{\tilde{x}_{0,\ell',i}^2}
\newcommand{\txbltisq}{\tilde{x}_{1,\ell',i}^2}
\newcommand{\axlisq}{\left|x_{\ell,i}^2\right|}
\newcommand{\awlisq}{\left|w_{\ell,i}^2\right|}

\newcommand{\dlfj}{d_{\ell,f}^{r}}
\newcommand{\slj}{s_{\ell}^{r}}

\newcommand{\dvxalj}{\dot{\mathbf{x}}_{0,\ell}^{j}}
\newcommand{\dvxblj}{\dot{\mathbf{x}}_{1, \ell}^{j}}
\newcommand{\vxalj}{\mathbf{x}_{0,\ell}^{j}}
\newcommand{\vxblj}{\mathbf{x}_{1, \ell}^{j}}
\newcommand{\cT}{c^{\textrm{ET}}}
\newcommand{\cTC}{c^{\textrm{TT}}}
\newcommand{\cM}{c^{\textrm{M}}}
\newcommand{\Lj}{L_{j}}
\newcommand{\cVj}{\mathbf{c}_{j}}
\newcommand{\sVj}{\mathbf{s}_{j}}
\newcommand{\Qj}{\mathbf{Q}_{j}}
\newcommand{\valpha}{\mathbf{\alpha}}
\newcommand{\nl}{n_l}
\newcommand{\nlt}{n_l'}
\newcommand{\ijp}{i_{j,p}}
\newcommand{\uijp}{\underline{i}_{j,p}}
\newcommand{\cVjp}{\mathbf{c}_{j,p}}
\newcommand{\vVl}{\mathbf{V}_{\ell}}
\newcommand{\djp}{d_{j,p}}
\newcommand{\vdj}{\mathbf{d}_{j}}
\newcommand{\vxal}{\mathbf{x}_{0,\ell}}
\newcommand{\vxbl}{\mathbf{x}_{1, \ell}}
\newcommand{\vxalT}{\mathbf{x}_{0,\ell}^{T}}
\newcommand{\vxblT}{\mathbf{x}_{1, \ell}^{T}}
\newcommand{\vzl}{\mathbf{z}_{\ell}}
\newcommand{\vwlT}{\mathbf{w}_{\ell}^{T}}
\newcommand{\vxlT}{\mathbf{x}_{\ell}^{T}}
\newcommand{\hvzl}{\hat{\mathbf{z}}_{\ell}}
\newcommand{\tvzl}{\tilde{\mathbf{z}}_{\ell}}
\newcommand{\tzlk}{\tilde{z}_{\ell,k}}
\newcommand{\azlk}{\left|{z}_{\ell,k}\right|}
\newcommand{\tzlksq}{\tilde{z}_{\ell,k}^{2}}
\newcommand{\azlksq}{\left|{z}_{\ell,k}\right|^{2}}
\newcommand{\tvzlj}{\tilde{\mathbf{z}}_{\ell}^{r}}
\newcommand{\dvzlj}{\dot{\mathbf{z}}_{\ell}^{r}}
\newcommand{\vzlj}{\mathbf{z}_{\ell}^{r}}
\newcommand{\alphaf}{\alpha_{f}}
\newcommand{\sensl}{s_{\ell}}
\newcommand{\deltaTf}{\delta_{\textrm{T},f}}
\newcommand{\deltaMf}{\delta_{\textrm{M},f}}
\newcommand{\clf}{c_{\ell,f}}
\newcommand{\vYtrue}{\mathbf{Y}_\textrm{true}}
\newcommand{\bbR}{\mathbb{R}}
\newcommand{\bbZ}{\mathbb{Z}}
\newcommand{\Dcalib}{\mathcal{D}_{\textrm{calib}}}
\newcommand{\Vertices}{\mathrm{Vertices}}
\newcommand{\Edges}{\mathrm{Edges}}
\newcommand{\vertexstart}{\mathrm{start\_vertex}}
\newcommand{\vertexend}{\mathrm{end\_vertex}}
\newcommand{\vertex}{\mathrm{vertex}}
\newcommand{\pathlen}{\mathrm{path\_len}}
\newcommand{\curlen}{\mathrm{cur\_len}}

\acrodef{MP}{Mixed-Precision}
\acrodef{NN}{Neural Network}
\acrodef{IP}{Integer Programming}
\acrodef{IP-TT}{IP-TheoreticalTime}
\acrodef{IP-ET}{IP-EmpiricalTime}
\acrodef{IP-M}{IP-Memory}
\acrodef{BGEMM}{Batch General Matrix Multiplication}
\acrodef{MAC}{Multiply and Accumulate}
\acrodef{MP}{Mixed Precision}
\acrodef{TTFT}{Time To First Token}
\acrodef{STE}{Straight-Through Estimator}
\acrodef{QAT}{Quantization-Aware Training}
\acrodef{QFT}{Quantization-Aware Fine-Tuning}
\acrodef{PTQ}{Post-Training Quantization}
\acrodef{HAWQ}{Hessian AWare Quantization}
\acrodef{OMPQ}{Orthogonal Mixed Precision Quantization}
\acrodef{DNAS}{Differentiable Neural Architecture Search}
\acrodef{HAQ}{Hardware-Aware Automated Quantization}
\acrodef{StruM}{Structured Mixed-precision}
\acrodef{ReLeQ}{Reinforcement Learning Approach for Deep Quantization}
\acrodef{MCKP}{Multiple-Choice Knapsack Problem}
\acrodef{LLM}{Large Language Model}
\acrodef{MSE}{Mean Square Error}
\acrodef{DAG}{Directed Acyclic Graph}
\acrodef{BFS}{Breadth-first search}
\acrodef{RMSE}{Root Mean Square Error}
\acrodef{1B}{Meta-Llama-3.2-1B-Instruct}
\acrodef{8B}{Meta-Llama-3.1-8B-Instruct}

\maketitle

\begin{abstract}
Quantization is essential for \ac{NN} compression, reducing model size and computational demands by using lower bit-width data types, though aggressive reduction often hampers accuracy. \ac{MP} mitigates this tradeoff by varying the numerical precision across network layers. This study focuses on automatically selecting an optimal \ac{MP} configuration within \ac{PTQ} for inference. The first key contribution is a novel sensitivity metric derived from a first-order Taylor series expansion of the loss function as a function of quantization errors in weights and activations. This metric, based on the \ac{MSE} of the loss, is efficiently calculated per layer using high-precision forward and backward passes over a small calibration dataset. The metric is additive across layers, with low calibration memory overhead as weight optimization is unnecessary. The second contribution is an accurate hardware-aware method for predicting \ac{MP} time gain by modeling it as additive for sequential sub-graphs. An algorithm partitions the model graph into sequential subgraphs, measuring time gain for each configuration using a few samples. After calibrating per-layer sensitivity and time gain, an \ac{IP} problem is formulated to maximize time gain while keeping loss \ac{MSE} below a set threshold. Memory gain and theoretical time gain based on \ac{MAC} operations are also considered. Rigorous experiments on the Intel Gaudi 2 accelerator validate the approach on several \acp{LLM}.


\acresetall

\end{abstract}

\vspace{-0.5em}
\section{Introduction}\label{sec:intro}
Quantization is a key technique for compressing neural networks by converting high-precision weights and activations to lower-precision formats, significantly reducing model size and computational load (\cite{rokh2023quantization}, \cite{gholami2021quantization}, \cite{nagel2021whitepaper}, \cite{guo2018quantization}, \cite{weng2021efficient}\cite{lee2025fasterinferencellmsusing}). This is crucial for efficient inference on accelerators and edge devices. The main quantization approaches are \ac{QAT}, \ac{PTQ}, and \ac{QFT}.

\ac{QAT} incorporates quantization during training, maintaining accuracy despite aggressive quantization (e.g., INT4) but requires substantial data and computational resources (\cite{jacob2018quantization}, \cite{krishnamoorthi2018quantizing}). In contrast, \ac{PTQ} quantizes pre-trained models efficiently but may reduce accuracy, especially in low-precision scenarios (\cite{migacz2017eight}, \cite{banner2019post}). \ac{QFT} combines \ac{PTQ} with fine-tuning to balance accuracy and resource efficiency (\cite{ashkboos2024efqat}).

\ac{MP} has emerged as a key technique for optimizing \ac{NN} performance across hardware platforms (\cite{rakka2024review}). Automatic selection of \ac{MP} configuration for inference involves search-based methods (e.g., \ac{HAWQ} and \ac{OMPQ}) and optimization-based approaches (e.g., \ac{DNAS}, \ac{HAQ}, \ac{ReLeQ}, AUTOQ). Optimization-based methods tackle the non-differentiability of bit-widths through reinforcement learning and hardware feedback.

\citet{pandey2023practical} proposed an \ac{MP} algorithm for post-training scenarios, minimizing data usage and considering hardware limitations. The algorithm first measures layer sensitivity, then reduces bit-width iteratively while maintaining performance. \citet{wu2025strumstructuredmixedprecision} introduced \ac{StruM}, tailored for compatible hardware, while \cite{chen2021mixedprecisionquantizationneuralnetworks} formulated loss minimization as a \ac{MCKP}, solved with a greedy search algorithm.

In this work, we address the challenge of selecting an \ac{MP} configuration that optimizes gained time or memory compared to the high-precision model for \ac{PTQ}, while maintaining a constraint on the \ac{MSE} of the loss. To solve this problem, we introduce a novel sensitivity metric derived from the \ac{MSE} of the loss of the quantized model. This metric is formulated by approximating the loss error as a first-order Taylor series expansion of quantization errors from weights and activations. It is estimated using both forward and backward passes of the model at high precision with a calibration dataset. The metric predicts the \ac{MSE} of the loss for arbitrary \ac{MP} configurations by considering the loss error components from different quantized layers as statistically independent. The \ac{MSE} of a given component from a layer is calculated as the product of its sensitivity and the \ac{MSE} of the quantization error of an individual element. Although the method requires a backward pass, the additional memory requirement is minimal, mainly consisting of stored activations (since weighta optimizer is not required).

We also introduce a method for predicting the empirical time gained from a \ac{MP} configuration, based on the additive execution time of sequentially computed sub-graphs. The model structure is analyzed to find sequential sub-graphs, each potentially comprising multiple layers or a single layer. Time gains for all \ac{MP} configurations are measured for each group, enabling the prediction of gained time for any configuration. The total time gain is estimated as the sum of the gained times per group.

The performance metric can also be expressed as memory gained or theoretical gained time based on the number of \ac{MAC} operations per layer. To optimize the metric while constraining the \ac{MSE} of the loss, we employ \ac{IP}.

The paper continues as follows: Sec.~\ref{sec:prop_method} presents the proposed method, Sec.~\ref{sec:exp} details experimental results, and Sec.~\ref{sec:conc} discusses conclusions. Experiments include results from applying the method on various \acp{LLM} and validating estimates for loss \ac{MSE} and gained time.

\vspace{-0.5em}
\section{Proposed method}\label{sec:prop_method}

In Sec.~\ref{subsec:IP} the problem is formulated and the solution is derived based on \ac{IP} optimizing a generic objective function with constrained loss \ac{MSE} per group (sequential sub-graph). Then, the sensitivity and the loss \ac{MSE} are derived in Sec.~\ref{subsec:sensitivity}. Various performance metrics which can substitute the generic objective function are defined in Sec.~\ref{subsec:computation}. We also discuss the motivation and method for partitioning the model graph to sequential sub-graphs for accurately assessing the time gained by \ac{MP}. The method is summarized in Sec.~\ref{sec:summary}.

\vspace{-0.5em}
\subsection{Formulation}\label{subsec:IP}
Let $\M$, $\vX$ and $\vY$ respectively denote a \ac{NN}, the input and output, such that:
\begin{align}
\vY \triangleq \M\left(\vX\right)
\end{align} and let  $g\left(\M\left(\vX\right), \vYtrue\right)$ denote the loss function
where $\vYtrue$ represents the \emph{ground-truth target} corresponding to $\vY$.
The \ac{NN} is composed of $L$  linear operations, including both standard linear layers and \ac{BGEMM} layers. Assume that the underlying hardware accelerator supports $F$ distinct numerical formats. A per-layer \ac{MP} configuration $\cI^{\textrm{layer}}$ is defined by a set of $L\times F$ binary indicators (one per each combination of layer and numerical format):
\begin{align}
\cI^{\textrm{layer}}\triangleq \left\{i_{\ell,f}^{\textrm{layer}} \in \left\{0, 1\right\}\right\}_{\ell\in\left[0, L-1\right],f\in\left[0,F-1\right]}\label{eq:ilf}
\end{align}
where $\ell$ indexes the layers and $f$ indexes the numerical formats. Each layer is assigned exactly one numerical format, enforcing the constraint $\sum_{f}i_{\ell,f}^{\textrm{layer}} = 1$.
The numerical formats are assumed to be various floating-point representations, differentiated by their mantissa bit widths, denoted $\mf$. 

Now, suppose the model is partitioned into $J$ disjoint groups of layers, $\left\{V_j\right\}_{j=0}^{J-1}$, such that layers within a group exhibit dependent performance characteristics, while different groups are independent. Define each group as the set of layer indices comprising it, i.e., $V_j \triangleq \left\{\ell_{j,0}, \ldots, \ell_{j,\Lj-1}\right\}$,
where $\Lj$ is the number of layers in group $j$. 
Let $\Qj\in\bbZ^{\Lj \times F^{\Lj}}$ be the matrix enumerating all per-layer possible quantization configurations for group $j$, where each column specifies a choice of numerical formats for the group layers, and each of its elements is in the range $\left[0,F-1\right]$.

We extend the standard per-layer binary indicator into a per-group binary indicator as follows. A per-group \ac{MP} configuration $\cI$, also denoted here as an \ac{MP} configuration for brevity, is defined by a set of $J\times F^{\Lj}$ binary indicators (one per each combination of group and any of its $F^{\Lj}$ possible quantization combinations):
\begin{align}
\cI\triangleq \left\{i_{j,p} \in \left\{0, 1\right\}\right\}_{j\in\left[0, J-1\right],p\in\left[0,F^{\Lj}-1\right]}\label{eq:ijp}
\end{align}
where $j$ indexes the groups and $p$ indexes its quantization configurations (indicating that the configuration in $p$-th column of $\Qj$ is selected). Each group is assigned exactly one configuration, enforcing the constraint $\sum_{p}i_{j,p} = 1$. 
A special case arises when the entire model is sequential; this corresponds to $J=L$ single layer groups with  $V_\ell=\left\{\ell\right\}$.

Let $\cVj\in\bbR^{F^{\Lj}}$ be the vector of performance metric values associated with the configurations in $\Qj$, and $\vdj\in\bbR^{F^{\Lj}}$ the corresponding loss \ac{MSE} values.
Define the \ac{MSE} of the loss function due to approximation under an \ac{MP} configuration as:
\begin{align}
\txtE\left[\tg^2\right]=\txtE\left[\left(\hg-g\right)^2\right]\label{eq:p}
\end{align}
where $\E\left[\bullet\right]$ denotes the expectation operator and $\hg$ is the perturbed loss under the \ac{MP} configuration. 

Let $c$ be a performance metric to be maximized. In this study, we evaluate several metrics: \emph{empirical time gain} denoted $\cT$, \emph{theoretical time gain} estimated from the number of \ac{MAC} operations denoted $\cTC$ and \emph{memory gain} from reduced model size denoted $\cM$.
Execution under an \ac{MP} configuration $\cI$ aims to improve performance, while potentially increasing the loss \ac{MSE}. 





Assuming a maximum allowable loss \ac{MSE} of $ \tau^2\txtE\left[g^2\right]$, for a parameter $\tau<1$ (which is the normalized-\ac{RMSE} threshold), our objective is to determine the optimal \ac{MP} configuration by solving:
\begin{align}
\left\{\ijp\right\}_{j,p} = \textrm{argmax}&_{\left\{\uijp\right\}_{j,p}}\sum_{j,p}{\uijp\cVjp}\nonumber\\
\textrm{s.t.:}&\sum_{j,p}{\uijp\djp}\leq\tau^2\txtE\left[g^2\right],\; \sum_{p}{\uijp}=1:\ \forall j,\; \uijp\in\left\{0, 1\right\}:\ \forall j,p.\label{eq:ip}
\end{align}
Define the \ac{IP} loss \ac{MSE} and performance metrics as:
\begin{center}
\begin{minipage}{0.4\textwidth}
\begin{align}
    d\triangleq\sum_{j,p}{\uijp\djp}\label{eq:d}
\end{align}
\end{minipage}
\begin{minipage}{0.10\textwidth}
\begin{center}
\end{center}
\end{minipage}
\begin{minipage}{0.4\textwidth}
\begin{align}
    c\triangleq \sum_{j,p}{\uijp\cVjp}.\label{eq:c}
\end{align}
\end{minipage}
\end{center}
In Sec.~\ref{subsec:sensitivity} and Sec.~\ref{subsec:computation}, we derive explicit expressions for the latter, respectively.




\vspace{-0.5em}
\subsection{Loss \ac{MSE} metric}\label{subsec:sensitivity}
The model comprises a set of standard linear layers, denoted by $\Llin$, and a set of \ac{BGEMM} layers, denoted by $\LBGEMM$. A linear layer $\ell\in\Llin$ is defined by the operation: 
\begin{align}
\vYl = \vXl \vWlT +\vone_{N\times 1}\vblT \label{eq:lin_layer}
\end{align}
where the dimensions of the matrices are as follows: $\vXl\in\bbR^{N\times\Cl}$, $\vWl\in\bbR^{\Kl\times\Cl}$, $\vYl\in\bbR^{N\times\Kl}$, and $\vbl\in\bbR^{\Kl\times1}$. Here, $N$ represents the number of input samples.

A \ac{BGEMM} layer $\ell'$ is defined as:
\begin{align}
\vYlt = \vXalt \otimes \vXblt \label{eq:bgemm_layer}
\end{align}
where $\vXalt$ and $\vXblt\in\bbR^{N\times \Clt}$, and the output $\vYlt\in\bbR^{N\times 1}$. The operator $\otimes$ is defined such that the $n$-th element of $\vYlt$ is computed by $Y_{\ell',n,0} \triangleq \left(\venT\vXalt\right)\left(\venT\vXblt\right)^{T}$
with  the selection vector $\ven\in\bbR^{N\times 1}$ defined as $\ven^{T} \triangleq \left[\mathbf{0}_{1\times n-1}, 1, \mathbf{0}_{1\times N-n}\right]$
and is used to extract the $n$-th row of a matrix.

Let $\vzl$ represent the extended input of layer $\ell$, obtained by vectorizing the possibly quantized inputs. It is defined as:
\begin{align}
\vzl \triangleq \left[\begin{array}{ll}
\left[\vxlT, \vwlT\right]^{T}&;\ell\in\Llin\\
\left[\vxalT, \vxblT\right]^{T}&;\ell\in\LBGEMM
\end{array}
\right].
\end{align}
Define the vectorized representations as:
\begin{center}
\begin{minipage}{0.4\textwidth}
\begin{subequations}
\begin{align}
\vxl \triangleq& \vect\left(\vXl\right)\\
\vwl \triangleq& \vect\left(\vWl\right)
\end{align} 
\end{subequations}
\end{minipage}
\begin{minipage}{0.10\textwidth}
\begin{center}

\end{center}
\end{minipage}
\begin{minipage}{0.4\textwidth}
\begin{subequations}
\begin{align}
\vxal \triangleq& \vect\left(\vXal\right)\\
\vxbl \triangleq& \vect\left(\vXbl\right)
\end{align} 
\end{subequations}
\end{minipage}
\end{center}
with dimensions $\vxl\in\bbR^{N\Cl\times 1}$, $\vwl\in\bbR^{\Cl\Kl\times 1}$ and $\vxal,\vxbl\in\bbR^{N\Cl\times 1}$.

We now respectively derive expressions for the noisy loss arising from model quantization and the quantized extended input:
\begin{center}
\begin{minipage}{0.4\textwidth}
\begin{align}
\hg \triangleq g+\tg, \label{eq:hg}
\end{align}
\end{minipage}
\begin{minipage}{0.10\textwidth}
\begin{center}

\end{center}
\end{minipage}
\begin{minipage}{0.4\textwidth}
\begin{align}
    \hvzl \triangleq \vzl+\tvzl
\end{align}
\end{minipage}
\end{center}
where $\tvzl$ is the quantization noise for layer $\ell\in\Llin\bigcup\LBGEMM$. 
And since $f$ represents a floating-point format with $\mf$ mantissa bits, the noise, modeled as a scaled Uniform random variable, and its respective variance are given by:
\begin{center}
\begin{minipage}{0.4\textwidth}
\begin{align}
    \tzlk \sim& \azlk 2^{-\mf} \textrm{U} [\pm 1/2]\label{eq:tzlk}
\end{align}
\end{minipage}
\begin{minipage}{0.10\textwidth}
\begin{center}
\end{center}
\end{minipage}
\begin{minipage}{0.4\textwidth}
\begin{align}
    \txtE\left[\tzlksq\right] =& \azlksq \alphaf\label{eq:var_tzlk}
\end{align}
\end{minipage}
\end{center}
for $k\in\left[0,\left|\vzl\right|\right]$, where $\textrm{U} [\pm 1/2]$ is a Uniform random distribution over $\left[-0.5, 0.5\right]$, and $\left|\vzl\right|$ denotes the number of elements in $\vzl$ with $\alphaf \triangleq \frac{2^{-2\mf}}{12}$ for $f\in\left[0, F-1\right]$.

Considering the $r$-th input sample and \eqref{eq:hg}, the noisy loss is expressed as $\hgj \triangleq \gj+\tgj$.
Assuming that the quantization noise is small compared to the full-precision values, a first-order Taylor series approximation yields:

\begin{center}
\begin{minipage}{0.40\textwidth}
\begin{equation}
\hgj \approx \gj+\sum_{\ell\in \Llin\bigcup\LBGEMM}\left(\tvzlj\right)^{T}\dvzlj
\end{equation}
\end{minipage}
\begin{minipage}{0.10\textwidth}
\begin{center}

\end{center}
\end{minipage}
\begin{minipage}{0.40\textwidth}
\begin{equation}
\dvzlj \triangleq \left.\frac{\partial g}{\partial \vzl}\right|_{\vzlj}
\end{equation}
\end{minipage}
\end{center}
where \(\dvzlj\) is the gradient of the loss with respect to the extended input $\vzl$ of sample $r$.

The \emph{sensitivity} of layer $\ell$ and its corresponding loss \ac{MSE} for numerical format $f$ and sample $r$ are respectively defined as: 
\begin{center}
\begin{minipage}{0.45\textwidth}
\begin{equation}
\slj \triangleq \|\vzlj\odot\dvzlj\|^2
\end{equation}
\end{minipage}
\hfill
\begin{minipage}{0.45\textwidth}
\begin{equation}
d_{\ell,f}^{\textrm{layer},r} \triangleq  \slj\alphaf. \label{eq:plfj}
\end{equation}
\end{minipage}
\end{center}
The variance of the contributions to the loss \ac{MSE} which correspond to the elements of the extended input are added in super-position, and with $\odot$ denoting the element-wise product.
Averaging over $R$ input samples yields the \emph{average} sensitivity and corresponding loss \ac{MSE} component:
\begin{center}
\begin{minipage}{0.45\textwidth}
\begin{equation}
\sensl \triangleq \frac{1}{R}\sum_{r}\slj\label{eq:sensl}
\end{equation}
\end{minipage}
\hfill
\begin{minipage}{0.45\textwidth}
\begin{equation}
d_{\ell,f}^{\textrm{layer}} \triangleq \sensl\alphaf.
\end{equation}
\end{minipage}
\end{center}
For the $p$-th quantization configuration, and under the assumption that quantization noise is statistically independent across layers, the loss \ac{MSE} component which corresponds to the $j$-th group is given by the sum of per-layer contributions:
\begin{align}
    d_{j,p} \triangleq \sum_{l=0}^{\Lj-1}s_{\ell_{j,l}} \alpha_{Q_{j,lp}}.
\end{align}

\vspace{-0.5em}
\subsection{Performance metric}\label{subsec:computation}
The choice of the performance metric $c$ significantly impacts the resulting \ac{MP} configuration. We consider three metrics: empirical time gain, theoretical time gain, and memory gain.

\subsubsection{Empirical Time Gain \texorpdfstring{$\cT$}{c\^T}}
\label{subsec:measured_time}

\textbf{Model partition to sequential sub-graphs:} 
\label{sec:model_partition}
The partition process is briefly described. For more details please refer to Sec. \ref{sec:model_sub_graph_algo}. Consider representing the computation of a model as a \ac{DAG}. Note that two adjacent sub-graphs that are connected by a single edge are computed sequentially since the second sub-graph \emph{depends} on the output of the first sub-graph. This sequential computation allows us to model their combined computation time as the sum of their individual times, which also applies to their gained time.
Our partition procedure identifies \emph{single-entry/single-exit sub-graphs} bounded by branching and merging nodes, splitting the computation graph into as many sequential sub-graphs as possible. These sub-graphs form an ordered sequence $\{V_j\}_{j=0}^{J-1}$ that executes strictly sequentially at run-time.
Predicting the computation time of concurrent layers within sub-graphs presents significant challenges. Operations within a sub-graph may execute in parallel, while the compiler is free to fuse or reorder operations. Additionally, latency depends on complex interactions between layer dependencies, hardware resources, and scheduling rules. We propose to avoid this complication and measure the gained time of each sub-graph, represented as a group of layers comprising it, for all their possible quantization configurations.

\begin{figure}[H]
  \centering
  \includegraphics[width=1\linewidth]{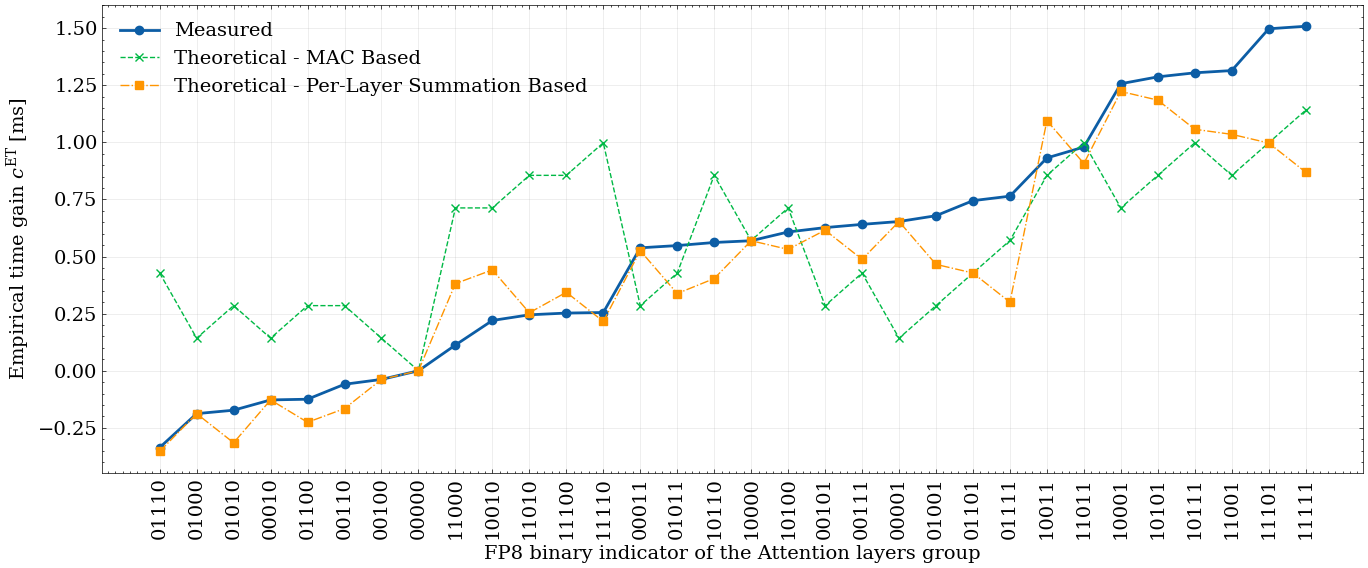}
  \caption{Measured empirical time gain $\cT_{j,p}$ of the Attention sub-graph in \textsc{Llama-3.1-8B} (in blue) compared to its prediction based on the summation of per-layer time gain measurements (in orange) and for the theoretical time gain $\cTC_{j,p}$ (green) for any of its $2^{5}$ \ac{MP} configurations. The various configurations are ordered in \textit{ascending order of  empirical time gain}. Configurations are labeled as 5-bit binary words which represent the numerical format of each of the 5 linear operations (         \texttt{q\_proj}, \texttt{v\_proj}, \texttt{k\_proj}, \texttt{qk\_matmul}, and \texttt{av\_matmul}) with BF16 and FP8 denoted as 0 and 1, respectively.}
  \label{fig:attentino_sub_graphs}
\end{figure}

\textbf{Gained time based on empirical time measurements per-group vs. per-layer:}\label{sec:gain_time_verif}
Consider the gained time of the Attention sub-graph in \textsc{Llama-3.1-8B}, illustrated in Figure~\ref{fig:sg_layer}, which contains the quantizable layers:  \texttt{q\_proj}, \texttt{v\_proj}, \texttt{k\_proj}, \texttt{qk\_matmul} and \texttt{av\_matmul}. Figure~\ref{fig:attentino_sub_graphs} compares the measured empirical
time gain $\cT_{j,p}$ of the attention sub-graph against the theoretical time gain predicted as the corresponding sum of per-layer time gain measurements. The large discrepancies demonstrate that simple summation of per-layer measurements does not yield a good estimate for the time gain of a sub-graph which contains concurrent computations. It shows the gap that the proposed method addresses.

\textbf{Gained time measurements:} The time gain of the $p$-th \ac{MP} configuration of the $j$-th group is measured by subtracting the end-to-end \ac{TTFT} of the model with the $j$-th group configured correspondingly and the other groups configured to BF16 from the end-to-end \ac{TTFT} of the model in BF16.



\subsubsection{Theoretical time gain \texorpdfstring{$\cTC$}{c\^TC}}
\label{subsec:theoretical_time}
This performance metric is defined per-layer as the theoretical time gain based on the number of \ac{MAC} operations multiplied by the gained time of a single \ac{MAC} in the $f$-th numerical format (compared to BF16), denoted $\deltaTf$.

For a linear layer
$\ell\in\{\Llin , \LBGEMM \}$ with $N$ samples, input dimension $\Cl$, and output
dimension $\Kl$ the theoretical time gain is defined as:
\begin{align}
\clf \triangleq \left\{\begin{array}{ll}
N \Cl \Kl \deltaTf &;\ell\in\Llin\\
N \Clsq \deltaTf&;\ell\in\LBGEMM
\end{array}
\right.
\end{align}
It is a simple performance metric that approximates the gained time without requiring any timing measurements. 

We compare theoretical and empirical time gains, i.e., $\cTC_{j,p}$ and $\cT_{j,p}$, for the Attention sub-graph in \textsc{Llama-3.1-8B}.
By definition, since the theoretical time gain is based on number of \acp{MAC}, it is additive across layers. Therefore, the theoretical time gain of the $p$-th configuration of the $j$-group is 
$\cTC_{j,p}=\sum_{l\in V_j}\cTC_{\ell,Q_{j,p}}$.
Figure~\ref{fig:attentino_sub_graphs} compares the aforementioned theoretical versus measured time gain. In order to simplify the comparison we fit the theoretical and empirical time gains, by constant scale and bias which we apply to the theoretical time gain. 
Even after optimal fitting, the theoretical proxy fails to
capture the measured behavior, indicating that MAC counts do not
reflect kernel fusion, memory traffic, or scheduler effects.  Note that the \ac{IP} is not affected by multiplying the performance metric by a scale factor and adding a bias to it.

\subsubsection{Memory Gain \texorpdfstring{$\cM$}{c\^M}}
\label{subsec:memory_gain}
\newcommand{\deltaM}{\Delta\!M}
Memory savings arise exclusively from storing weights at lower
precision.  
Intermediate tensors produced by \ac{BGEMM} kernels can certainly be
computed in \textsc{FP8}, but since they are not persistent they are stored in the stack memory. Quantizing them therefore improves latency but \emph{does not} change the static model size.
Under these observations, the additivity assumption across layers holds.
Let $\delta_{M,f}$ be the byte reduction obtained when a single parameter
element is stored in format $f$ instead of \textsc{BF16}.


Since memory is additive across layers, we treat each primitive layer as
its own group, i.e.\ $J=L$ and $V_j=\{\ell_j\}$.
For a (trivial) group $j$ and bit-width assignment $Q_{j,p}$ the memory gain is:


\begin{center}
\begin{minipage}{0.49\textwidth}
\begin{equation}
\label{eq:mem_per_layer}
c_{\ell,f} \;\triangleq\;
\begin{cases}
  \Cl\Kl\delta_{M,f} & \ell \in \Llin, \\[4pt]
  0                    & \ell \in \LBGEMM .
\end{cases}
\end{equation}
\end{minipage}
\hfill
\begin{minipage}{0.49\textwidth}
\begin{equation}
\cVjp \triangleq \sum_{l=0}^{\Lj-1}c_{\ell_{j,l}, Q_{j,p}}.
\end{equation}
\end{minipage}
\end{center}

These $\cVjp$ values are used by the \ac{IP} with the objective of maximizing memory gain~$\cM$.


\vspace{-0.5em}
\section{Experimental results}\label{sec:exp}
This section is organized as follows. In Sec.~\ref{subsec:setup} the experiments setup as well as the compared strategies are defined. Using \ac{IP} to solve the accuracy vs. performance tradeoff optimally assumes that the loss \ac{MSE} model that we use is sufficiently accurate, that it is additive across layers and that the empirical time gain is additive across groups. We validate these assumptions in Sec.~\ref{subsec:model_validation}. The \ac{IP} in our proposed method applies a constraint on the loss \ac{MSE} instead of on the accuracy degradation since we assume that they are correlated and since the latter is non-differentiable, nor additive across groups. In Sec.~\ref{subsec:loss_mse_gain_time_curve} we analyze the loss \ac{MSE} vs. empirical time gain curve which is optimized by \ac{IP}. Later in Sec.~\ref{subsec:acc_perf} we analyze the accuracy vs. performance curve.


\subsection{Setup and compared \ac{MP} strategies}\label{subsec:setup}
We evaluate \ac{MP} quantization during the prefill stage of \ac{LLM} inference using Intel's Gaudi 2 accelerator with $F=2$ numerical formats (BF16 and FP8-E4M3: 4 exponent, 3 mantissa bits), the lm-evaluation-harness \cite{lmevalharness}, and Neural Compressor \cite{intelneuralcompressor}. Our evaluation spans four tasks (HellaSwag~\cite{zellers2019hellaswag}, LAMBADA~\cite{radford2019language}, Winogrande~\cite{ai2:winogrande}, and PIQA~\cite{Bisk2020}), averaging $5$ iterations per configuration for time measurement, 20\% of the samples in each dataset for calibration and sensitivity measurements, and the full datasets for final evaluation. Results are reported for \ac{1B} (with batch size 40) and \ac{8B} models (with batch size 10). Each evaluation is run over $10$ different randomization seeds in which we perturb the scales before quantization in order to assess the accuracy statistics (mean and standard-deviation) and not just a single noisy realization of it.

Our proposed method combined with the different metrics yields the following strategies: \ac{IP-ET} maximizes empirical time gain (Sec.~\ref{subsec:measured_time}), \ac{IP-TT} maximizes theoretical time gain (Sec.~\ref{subsec:theoretical_time}), and \ac{IP-M} maximizes memory gain (Sec.~\ref{subsec:memory_gain}). Both \ac{IP-ET} and \ac{IP-TT} quantize linear and \ac{BGEMM} layers , while \ac{IP-M} quantizes only linear layers. 

\begin{figure}[t]
    \centering
    \includegraphics[width=0.89\linewidth]{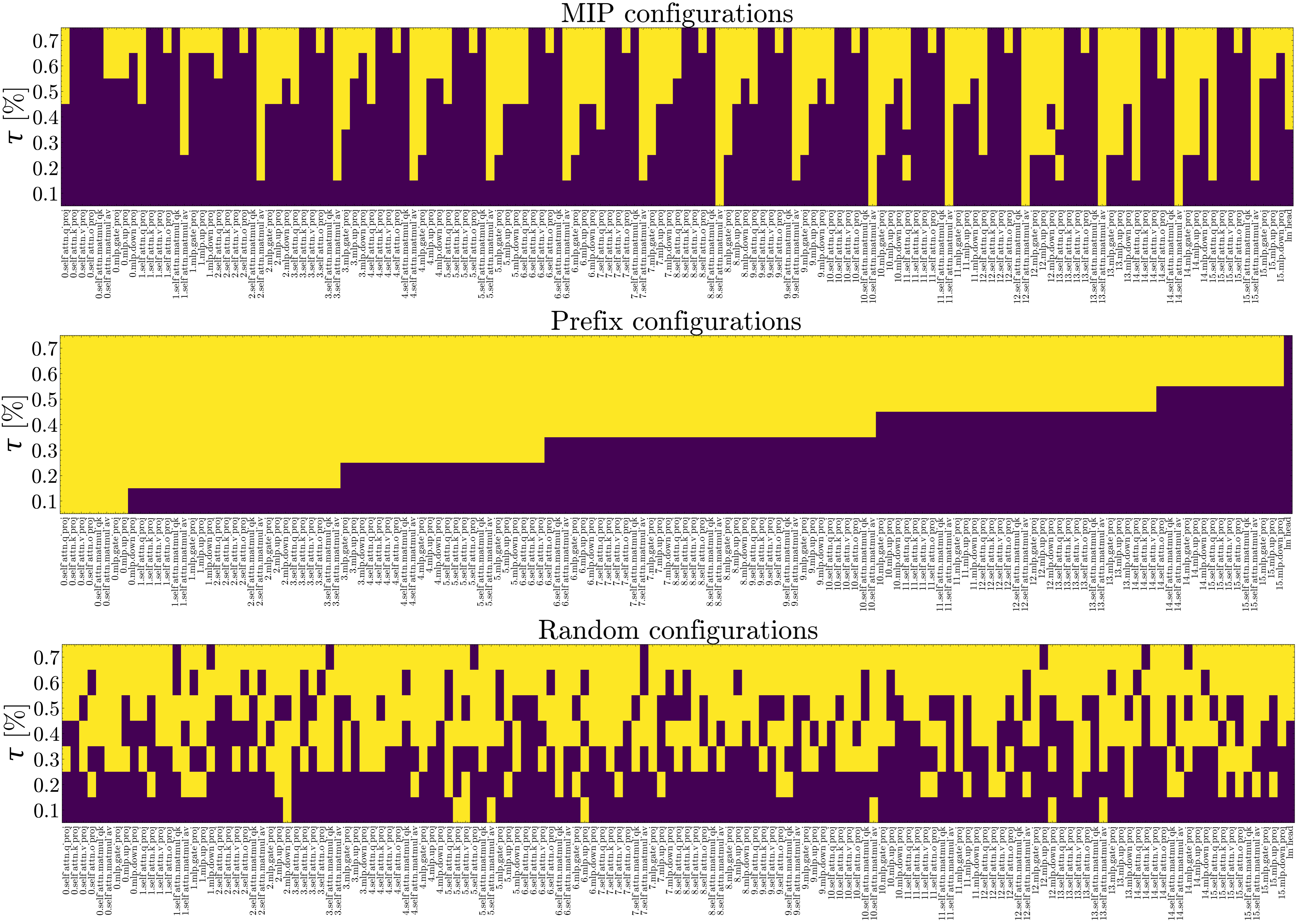}
    \caption{Layer-wise quantization patterns across \ac{MP} configurations (rows) and model layers (columns) for \ac{IP-ET} (top), Prefix (middle), and Random (bottom). Yellow: FP8, purple: BF16.}
    \label{fig:configs_visualization_hellaswag}
\end{figure}

Each of the \ac{IP} strategies is compared against two baseline strategies: \emph{Random} which arbitrarily selects layers to quantize, resulting in scattered patterns and \emph{Prefix} which quantizes layers in a sequential order. Both baseline strategies adhere to the loss \ac{MSE} threshold.
Figure~\ref{fig:configs_visualization_hellaswag} illustrates how each strategy selects layers for quantization given normalized-\ac{RMSE} threshold $\tau$. 
Our proposed \ac{IP-ET} strategy produces \emph{optimal} configurations which maximize the performance metric under the loss \ac{MSE} constraint, leading to its superior accuracy-performance curve shown in subsequent results.

\begin{figure}[t]
    \centering
    \begin{subfigure}{0.47\textwidth}
        \centering
        \includegraphics[width=\linewidth]{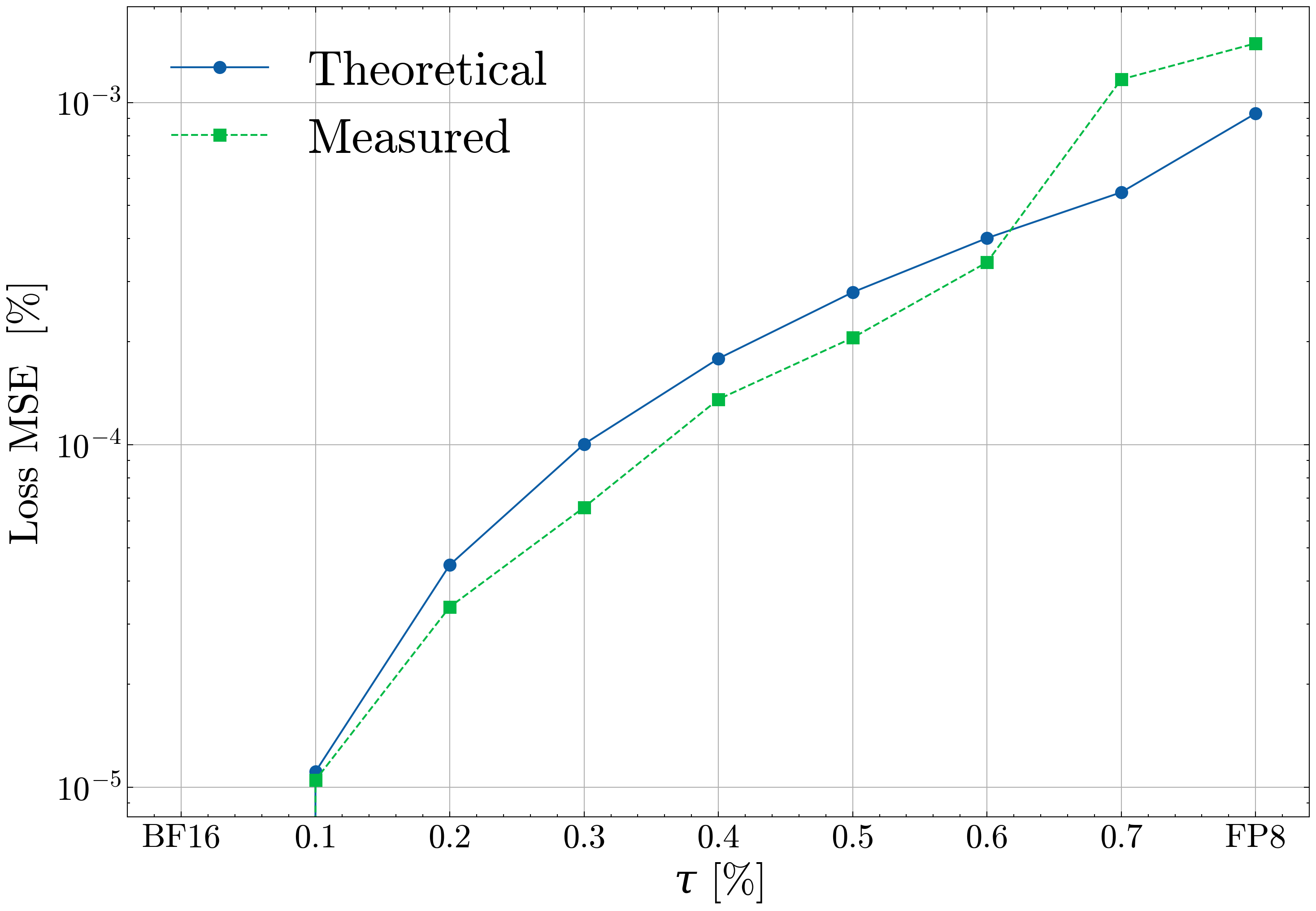}
        \caption{Loss MSE versus \(\tau\).
            Blue - theoretical loss MSE;
           Green - measured loss MSE using the chosen configurations by \ac{IP-ET}
           }
        \label{fig:Measured/meas_and_pred_sens_vs_ttft(IP)}
    \end{subfigure}
    \hfill
    \begin{subfigure}{0.47\textwidth}
        \centering
        \includegraphics[width=\linewidth]{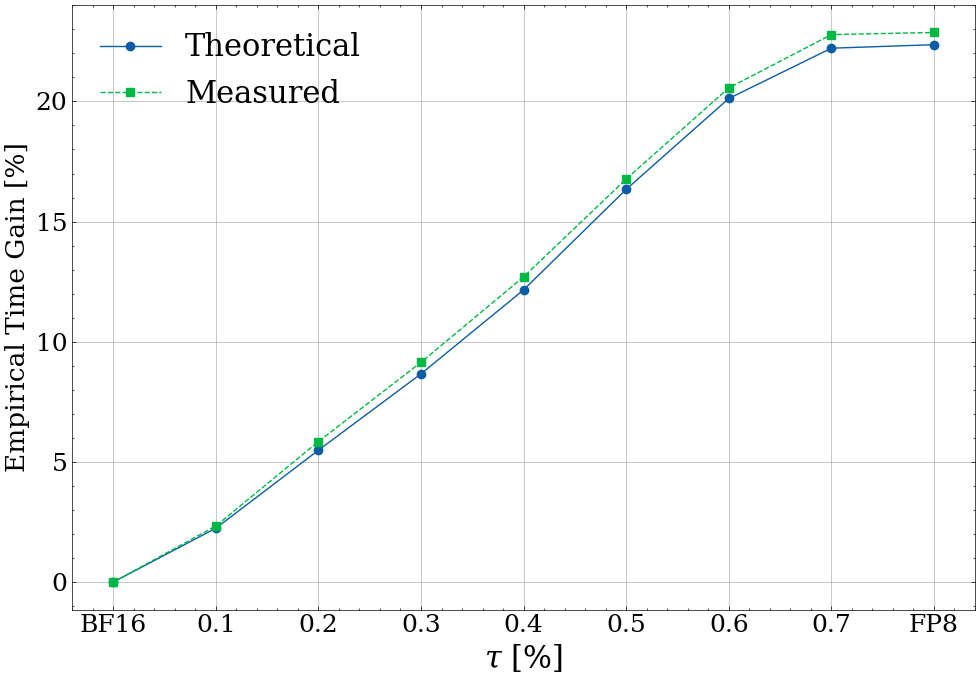}
        \caption{Relative \ac{TTFT} reduction versus \(\tau\).
           Blue - theoretical gain from the group aware \ac{IP-ET};  
           Green - measured gain on Gaudi 2.  
           }
        \label{fig:Performance/predicted_vs_measured}
    \end{subfigure}
    \caption{Empirical validation of the additivity assumption on different MP configurations}
    \label{fig:measured-predicted-comparison}
\end{figure}

\subsection{Time gain and loss \ac{MSE} model validation}\label{subsec:model_validation}
Considering the 1B model for \ac{MP} configurations attained using the proposed method with $\tau\in\{0,0.1\%,\dots,0.7\%\}$ in addition to the all-\textsc{FP8} configuration, we depict the measured vs. theoretical empirical time gain (see \eqref{eq:c}) and loss \ac{MSE}, respectively, in
Figure~\ref{fig:Measured/meas_and_pred_sens_vs_ttft(IP)} and Figure~\ref{fig:Performance/predicted_vs_measured}. Evidently our assumptions hold as the empirical time gain appears additive across groups and the theoretical loss \ac{MSE}, assuming the per-layer model \eqref{eq:sensl} and additivity \eqref{eq:d}, is a reliable estimate for the measured loss \ac{MSE}. 

\subsection{Loss \ac{MSE} vs. empirical time gain curve}\label{subsec:loss_mse_gain_time_curve}

\begin{figure}[htbp]
    \centering
    \includegraphics[width=0.5\linewidth]{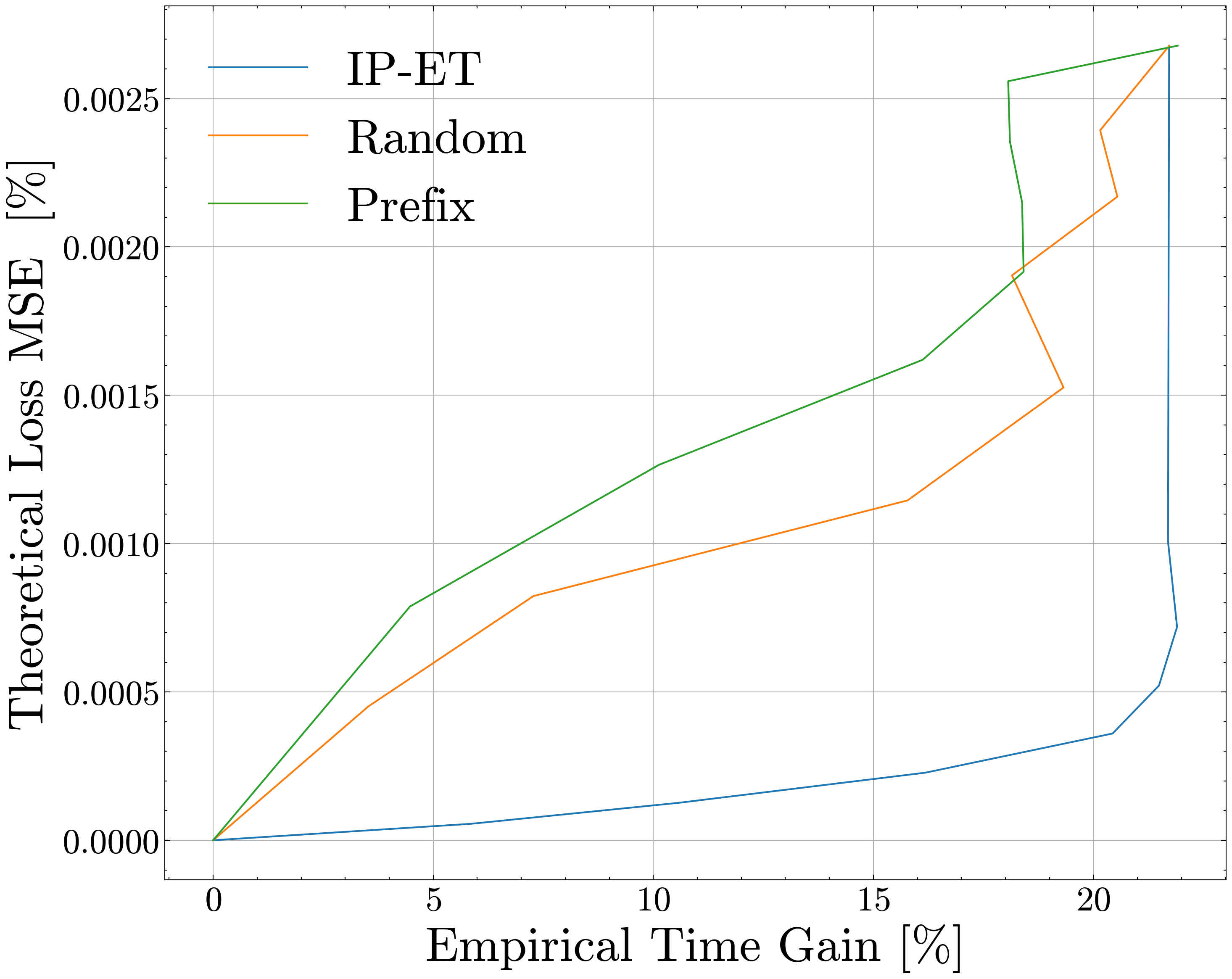}
    \caption{Theoretical loss \ac{MSE} vs. empirical time gain on the \ac{1B} model across four tasks.}
    \label{fig:Measured/pred_sens_vs_ttft}
\end{figure}

Figure \ref{fig:Measured/pred_sens_vs_ttft} demonstrates that the \ac{IP-ET} strategy is significantly and consistently better then the Random and Prefix strategies, yielding an appealing loss \ac{MSE} vs. empirical time gain curve. Furthermore, it maintains markedly low loss \ac{MSE} vs.  empirical time gain. 

\vspace{-1em}
\subsection{Accuracy vs. performance curve}\label{subsec:acc_perf}

\begin{figure}[H]
    \centering
    \begin{subfigure}{0.47\textwidth}
        \centering
        \includegraphics[width=\linewidth]{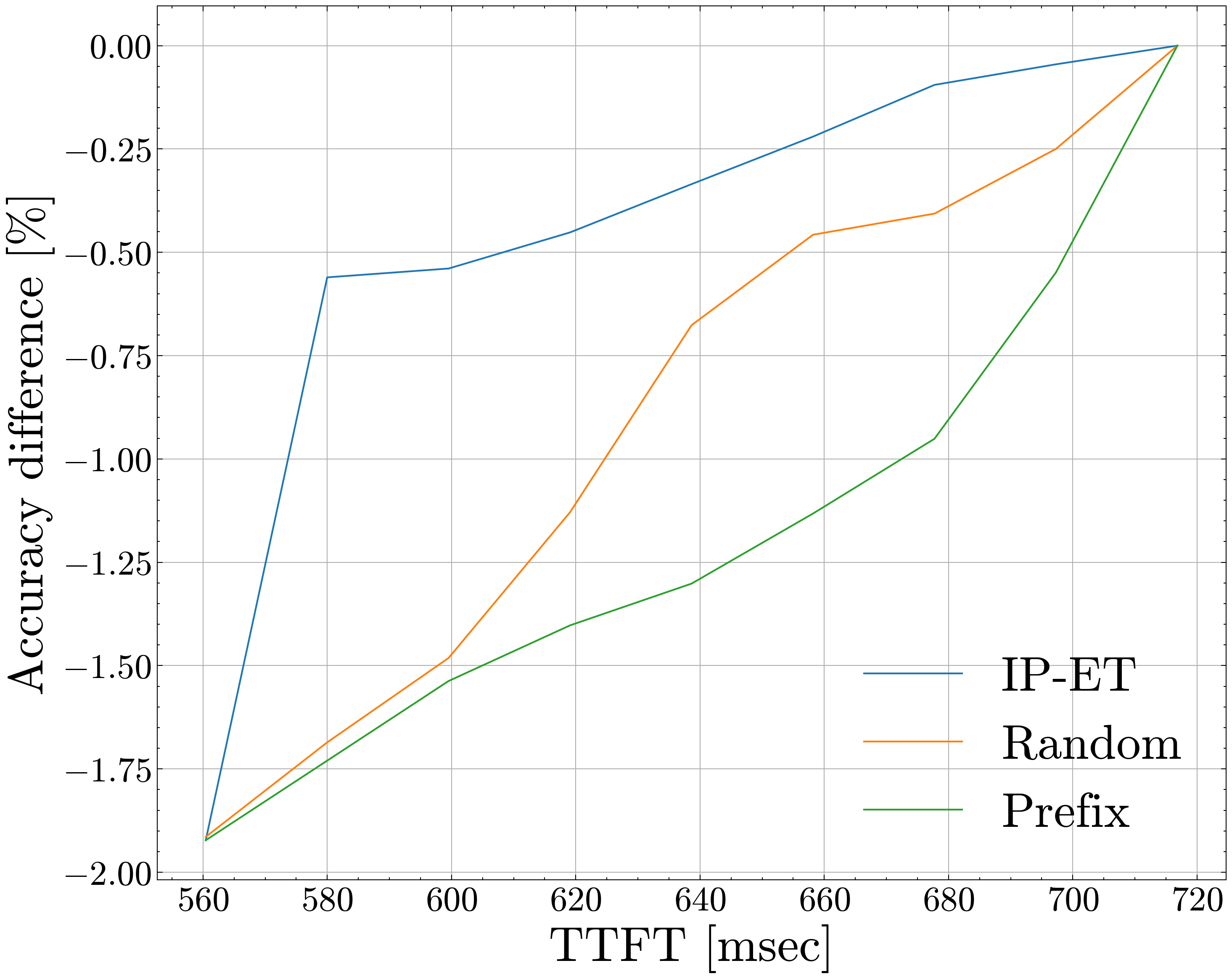}
        \caption{\ac{1B}}
        \label{fig:1B/Measured/acc_avg_vs_ttft}
    \end{subfigure}
    \hfill
    \begin{subfigure}{0.47\textwidth}
        \centering
        \includegraphics[width=\linewidth]{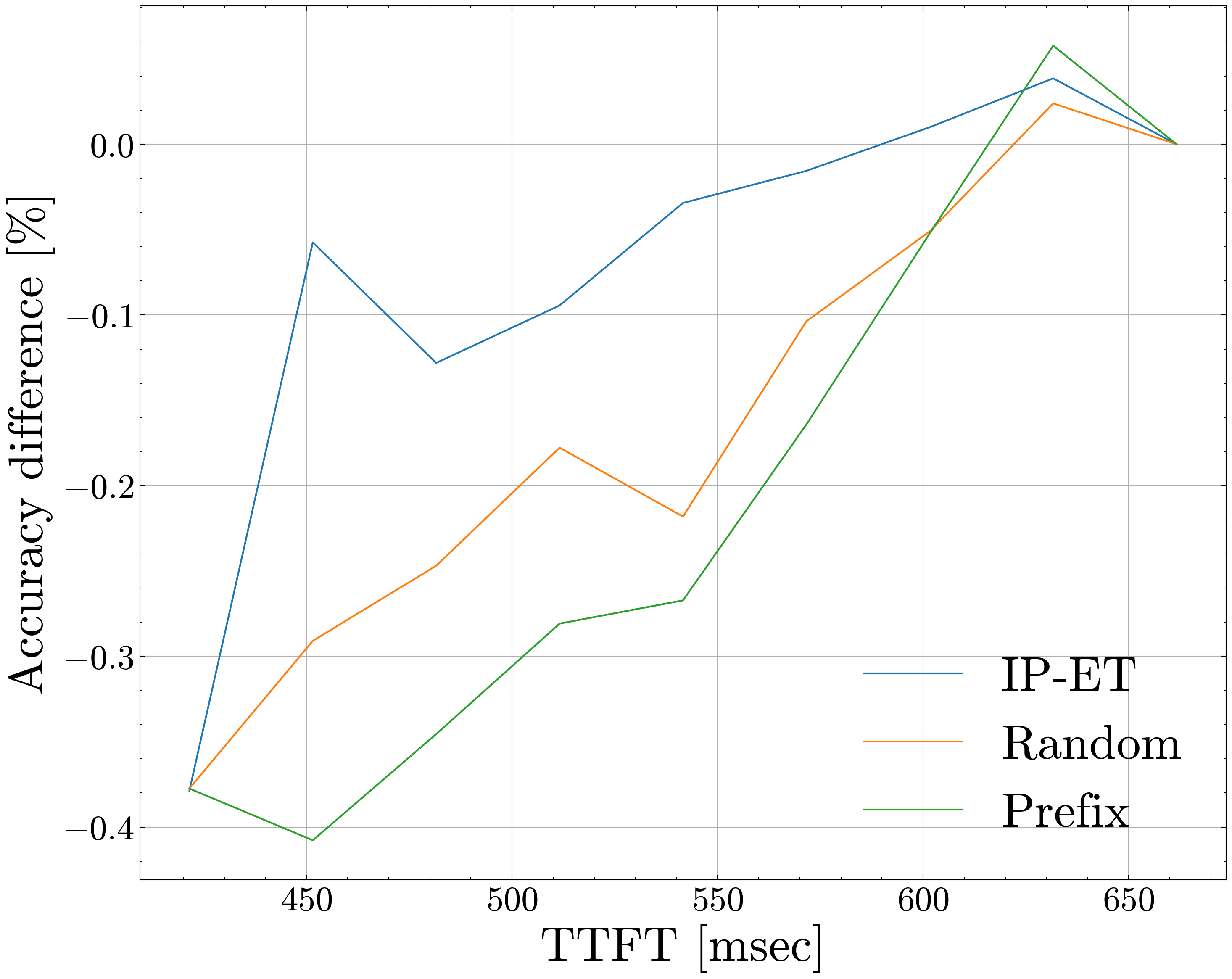}
        \caption{\ac{8B}}
        \label{fig:8B/Measured/acc_avg_vs_ttft}
    \end{subfigure}
    \caption{Average accuracy difference [\%] vs. \ac{TTFT} across HellaSwag, LAMBADA, Winogrande, and PIQA. Comparing \ac{MP} quantization strategies (\ac{IP-ET}, Random and Prefix)}
    \label{fig:ttft-comparison}
\end{figure}

Figure~\ref{fig:1B/Measured/acc_avg_vs_ttft} and Figure~\ref{fig:8B/Measured/acc_avg_vs_ttft} illustrate the accuracy degradation vs. \ac{TTFT} curves of different strategies for the 1B and 8B models, respectively. For both models, \ac{IP-ET} consistently achieves better accuracy at comparable latency than the baselines. E.g., with \ac{8B}, \ac{IP-ET} achieves accuracy loss below 0.1\% at 450ms \ac{TTFT}, whereas other strategies require $\sim$600ms for similar accuracy—a 30\% speedup.

Table~\ref{tab:merged_configs_avg} provides a comprehensive comparison across all strategies and models. The proposed \ac{IP}-based methods consistently outperform the baselines. Despite its limited quantization scope (linear layers only), \ac{IP-M} still surpasses the baselines in most cases, with one exception: for \ac{8B} on LAMBADA, the Prefix strategy achieves slightly higher accuracy. These results confirm that sensitivity-aware, hardware-informed quantization significantly improves inference efficiency while preserving model quality. The improvement of the proposed method for the \ac{1B} model is better then the \ac{8B} model since the gap between the 
FP8 and 
BF16 accuracies there is larger. See Sec.~\ref{sec:additional_results} for additional results on per-task time gains, MAC-based gains, and memory gains.

\vspace{-5pt}

\begin{table}[htbp]
\centering
\resizebox{\textwidth}{!}{%
\begin{tabular}{llcccccc}
\toprule
\textbf{Model} 
& \textbf{Strategy} 
& \textbf{LAMBADA} 
& \textbf{LAMBADA} 
& \textbf{HellaSwag} 
& \textbf{Winogrande} 
& \textbf{PIQA} 
& \textbf{Tasks Avg.} \\
\midrule
& & \textit{ppl diff \(\downarrow\) [\%]} 
  & \textit{acc diff \(\uparrow\) [\%]} 
  & \textit{acc diff \(\uparrow\) [\%]} 
  & \textit{acc diff \(\uparrow\) [\%]} 
  & \textit{acc diff \(\uparrow\) [\%]} 
  & \textit{acc diff \(\uparrow\) [\%]} \\
\midrule
\midrule[\heavyrulewidth]
\multicolumn{8}{c}{\textbf{IP-ET - Empirical Time Gain Optimization (both BGEMMs and linear layers)}} \\
\midrule[\heavyrulewidth]
\multirow{3}{*}{Llama-3.2-1B-Instruct}
    & Random                 &  4.938 $\pm$ 0.96&  -2.107 $\pm$ 0.45&  -1.077 $\pm$ 0.35&  \textbf{0.077 $\pm$ 0.93}&  -0.449 $\pm$ 0.34&  -0.889 $\pm$ 0.52\\
    & Prefix                 &  5.986 $\pm$ 1.61&  -2.206 $\pm$ 0.51&  -1.586 $\pm$ 0.43&  -0.271 $\pm$ 0.78&  -0.615 $\pm$ 0.55&  -1.170 $\pm$ 0.57\\
    & \ac{IP-ET} &  \textbf{2.170 $\pm$ 0.32}&  -\textbf{1.401 $\pm$ 0.26}&  \textbf{-0.303 $\pm$ 0.14}&  0.020 $\pm$ 0.59&  \textbf{-0.169 $\pm$ 0.21}&  \textbf{-0.463 $\pm$ 0.30}\\
\midrule
\multirow{3}{*}{Llama-3.1-8B-Instruct}
    & Random                 &  1.290 $\pm$ 0.15&  -0.256 $\pm$ 0.25&  -0.071 $\pm$ 0.08&  0.085 $\pm$ 0.55&  -0.399 $\pm$ 0.26&  -0.160 $\pm$ 0.286\\
    & Prefix                 &  1.075 $\pm$ 0.15&  -0.029 $\pm$ 0.24&  -0.157 $\pm$ 0.12&  -0.065 $\pm$ 0.66&  -0.566 $\pm$  0.30&  -0.204 $\pm$ 0.33\\
    & \ac{IP-ET} &  \textbf{0.922 $\pm$ 0.08}&  \textbf{-0.229 $\pm$ 0.17}&  	$\bm{2.53\mathrm{e}{-4} \pm 0.06}$&  \textbf{0.276 $\pm$ 0.41}&  \textbf{-0.341 $\pm$ 0.17}&  \textbf{-0.073 $\pm$ 0.20}\\
\midrule[\heavyrulewidth]
\multicolumn{8}{c}{\textbf{IP-TT - Theoretical Time Gain Optimization (both BGEMMs and linear layers)}} \\
\midrule[\heavyrulewidth]
\multirow{3}{*}{Llama-3.2-1B-Instruct}
    & Random                 &  4.938 $\pm$ 0.98&  -2.107 $\pm$ 0.46&  -1.077 $\pm$ 0.35&  0.077 $\pm$ 0.85&  -0.449 $\pm$ 0.33&  -0.889 $\pm$ 0.49\\
    & Prefix                 &  5.986 $\pm$ 1.62&  -2.206 $\pm$ 0.50&  -1.586 $\pm$ 0.43&  -0.271 $\pm$ 0.79&  -0.615 $\pm$ 0.54&  -1.170 $\pm$ 0.57\\
    & \ac{IP-TT}             &  \textbf{2.744 $\pm$ 0.43}&  \textbf{-1.697 $\pm$ 0.41}&  \textbf{-0.429 $\pm$ 0.14}&  \textbf{0.096 $\pm$ 0.61}&  \textbf{-0.102 $\pm$ 0.26}&  \textbf{-0.533 $\pm$ 0.35}\\
\midrule
\multirow{3}{*}{Llama-3.1-8B-Instruct}
    & Random                 &  1.290 $\pm$ 0.15&  -0.256 $\pm$ 0.25&  -0.071 $\pm$ 0.08&  0.085 $\pm$ 0.55&  -0.399 $\pm$ 0.26&  -0.160 $\pm$ 0.28\\
    & Prefix                 &  1.075 $\pm$ 0.15&  -0.029 $\pm$ 0.24&  -0.157 $\pm$ 0.12&  -0.065 $\pm$ 0.67&  -0.566 $\pm$ 0.31&  -0.204 $\pm$ 0.33\\
    & \ac{IP-TT}             &  \textbf{1.002 $\pm$ 0.08}&  \textbf{-0.178 $\pm$ 0.15}&  $\bm{2.58\mathrm{e}{-4} \pm 0.06}$&  \textbf{0.185 $\pm$ 0.43}&  \textbf{-0.279 $\pm$ 0.19}&  \textbf{-0.068 $\pm$ 0.21}\\
\midrule[\heavyrulewidth]
\multicolumn{8}{c}{\textbf{IP-M - Memory Gain Optimization (only linear layers)}} \\
\midrule[\heavyrulewidth]
\multirow{3}{*}{Llama-3.2-1B-Instruct}
    & Random       &  4.151 $\pm$ 1.25&  -1.886 $\pm$ 0.52&  -0.980 $\pm$ 0.35&  0.396 $\pm$ 0.87&  -0.363 $\pm$ 0.30&  -0.708 $\pm$ 0.51\\
    & Prefix       &  4.483 $\pm$ 1.41&  -1.693 $\pm$ 0.64&  -1.361 $\pm$ 0.41&  \textbf{0.435 $\pm$ 0.86}&  -0.554 $\pm$ 0.44&  -0.794 $\pm$ 0.59\\
    & \ac{IP-M} &  \textbf{2.497 $\pm$ 0.34}&  \textbf{-1.512 $\pm$ 0.33}&  \textbf{-0.421 $\pm$ 0.15}&  0.230 $\pm$ 0.67&  \textbf{-0.075 $\pm$ 0.26}&  \textbf{-0.445 $\pm$ 0.35}\\
\midrule
\multirow{3}{*}{Llama-3.1-8B-Instruct}
    & Random       &  1.073 $\pm$ 0.10&  -0.267 $\pm$ 0.21&  -0.024 $\pm$ 0.08&  0.180 $\pm$ 0.49&  -0.321 $\pm$ 0.24&  -0.108 $\pm$ 0.25\\
    & Prefix       &  \textbf{0.567 $\pm$ 0.13}&  \textbf{0.015 $\pm$ 0.18}&  -0.092 $\pm$ 0.07&  0.271 $\pm$ 0.47&  -0.457 $\pm$ 0.22&  -0.066 $\pm$ 0.23\\
    & \ac{IP-M} &  0.981 $\pm$ 0.08&  -0.160 $\pm$ 0.17&  \textbf{0.012 $\pm$ 0.06}&  \textbf{0.280 $\pm$ 0.37}&  -\textbf{0.262 $\pm$ 0.16}&  \textbf{-0.032 $\pm$ 0.19}\\
\bottomrule
\end{tabular}
}
\caption{Accuracy and perplexity difference across three optimization strategies, averaged over different quantization configurations from high-precision (BF16) to low-precision (FP8).}
\label{tab:merged_configs_avg}
\end{table}

\vspace{-5pt}

\vspace{-1em}
\section{Conclusions}\label{sec:conc}
\vspace{-0.5em}
By utilizing a novel \emph{loss \ac{MSE}} and \emph{empirical time gain per sequential sub-graphs} metrics, we introduce an automatic \ac{MP} method based on \ac{IP} for \ac{PTQ}. The proposed loss \ac{MSE} metric, which exhibits additive properties per layer, serves as a proxy for model accuracy. We efficiently approximate this metric using forward- and backward-passes over a small calibration dataset. Recognizing that the empirical time gain exhibits additivity solely for sequential sub-graphs—attributable to parallel capabilities and advanced compiler optimizations in the hardware accelerator— we formulate an algorithm for model partitioning. In this approach, each sub-graph is characterized as a group of constituent layers, and we define a performance objective function by summing the empirical time gain for each group. To achieve this, we measure the empirical time gains of each sub-graph over a limited set of samples. We validate both the approximation of the loss \ac{MSE} and its additive nature across layers. Furthermore, we demonstrate that the empirical time gain is additive per group, resulting in a highly accurate estimate of the measured time gain. Finally, we evaluate the proposed method by comparing it against baseline strategies (Random and Prefix configurations), demonstrating that it consistently outperforms these approaches across various \acp{LLM}.


\bibliographystyle{plainnat}
\bibliography{refs.bib}

\appendix
\section*{Appendix}
\section{Proposed method summary}\label{sec:summary}
Algorithm \ref{alg:mp_summary} summarizes our end-to-end approach for automatic \ac{MP} configuration. The method integrates hardware-aware timing measurements with gradient-based sensitivity analysis to determine optimal precision assignments. After partitioning the model into sequential sub-graphs (line 1), we perform sensitivity calibration through forward and backward passes on the calibration dataset (line 2). We then measure empirical time gains for each sub-graph across different precision configurations (line 3), before formulating and solving the \ac{IP} optimization that maximizes performance while respecting the loss \ac{MSE} threshold (line 4). This algorithm forms the foundation for all three optimization strategies (\ac{IP-ET}, \ac{IP-TT}, and \ac{IP-M}).
\begin{algorithm}[H]
\caption{Proposed automatic \ac{MP} algorithm summary}\label{alg:mp_summary}
\begin{algorithmic}[1]
\REQUIRE A model $\M$, a calibration dataset $\Dcalib$ and a relative \ac{RMSE} threshold $\tau$
\ENSURE \ac{MP} configuration $\cI$ (see \eqref{eq:ijp})
\STATE Analyze model and partition it to $J$ sequential sub-graphs $\left\{V_j\right\}_j$ as described in \ref{sec:model_sub_graph_algo}
\STATE Sensitivity calibration
\begin{itemize}
\item Wrap the model $\cM$ to enable sensitivity measurement
\item Run forward- and backward- passes over $\Dcalib$, and obtain: sensitivity $\left\{\sensl\right\}_{\ell}$ and mean-square loss $\txtE\left[g^2\right]$ (see \eqref{eq:sensl})
\end{itemize}
\STATE Empirical time gain measurement
\begin{itemize}
\item Measure \ac{TTFT} of $j$-th group and $p$-th \ac{MP} configuration, for $j\in\left[0,J-1\right]$ and $p\in\left[0, F^{\Lj}-1\right]$
\item Compute $\cT$ by subtracting the measurements from the \ac{TTFT} of the model in BF16, 
\end{itemize}
\STATE Obtain $\cI$ by solving the \ac{IP} optimization problem (see \eqref{eq:ip})
\RETURN $\cI$
\end{algorithmic}
\end{algorithm}

\section{Model Partitioning into sequential groups of layers}
\label{sec:model_sub_graph_algo}
\begin{figure}[t]
  \centering
  \subfloat[Attention block]{%
    \includegraphics[width=.89\linewidth]{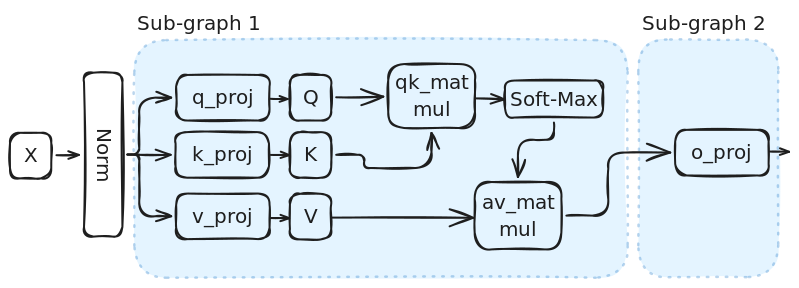}\label{fig:sg_attn}}
  \vspace{2mm}
  
  \subfloat[MLP block]{%
    \includegraphics[width=.6\linewidth]{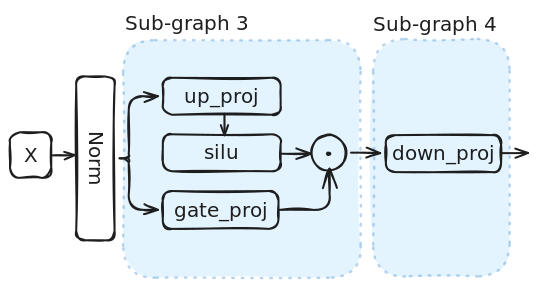}\label{fig:sg_mlp}}
  \caption{Single-entry/single-exit sub-graphs ($V_1$–$V_4$) identified
           in one Llama-3 transformer layer.  Dashed blue regions denote
           the latency-additive sub-graphs used in \ref{sec:gain_time_verif}; residual adds
           are omitted for clarity. The final LM-head forms an additional single-layer sub-graph that is omitted from the illustration for brevity.}
  \label{fig:sg_layer}
\end{figure}

Effective \ac{MP} assignment requires identifying model sub-graphs which execution time is additive. Given a network’s computation graph, that can be formulated as a \ac{DAG} with a single sink vertex - our partitioning algorithm splits the model to sequential sub-graphs with a single entry and a single exit points. Figure \ref{fig:sg_layer} illustrates the resulting partitioning for a Llama-3 transformer layer, showing the Attention and MLP blocks split into single-entry/single-exit sub-graphs ($V_1$–$V_4$) that serve as the fundamental units for our \ac{MP} optimization.

\begin{algorithm}
\caption{Partition model to sequential groups of layers}
\begin{algorithmic}[1]
\REQUIRE A model $\M$
\ENSURE Model partition $\left\{V_j\right\}_j$
\STATE Construct a \ac{DAG} graph of the model computation $\left\{\Vertices, \Edges\right\}$
\STATE Add a start vertex $\vertexstart$ and denote the end vertex as $\vertexend$
\STATE Run \ac{BFS} and denote the longest path from $\vertexstart$ to $\vertex$ as $\pathlen\left[\vertex\right]$ for each $\vertex\in\Vertices$
\STATE $V=[], \vertex=\vertexstart$
\WHILE{$\vertex\neq\vertexend$}
    \STATE Define set $V'=\{\}$
    \STATE $\curlen=\pathlen[\vertex]+1$
    \STATE Define the set $A=\mathrm{next}[\vertex]$
    \WHILE{$|A|>1$}
        \FOR{$\vertex'\in A$}
            \IF{$\pathlen[\vertex']\leq \curlen$}
                \STATE $A.\mathrm{pop}(\vertex')$
                \STATE $V'.\mathrm{push}(\vertex')$
                \STATE $A.\mathrm{push}(\mathrm{next}[\vertex'])$
            \ENDIF
        \ENDFOR
        \STATE $\curlen=\curlen+1$
    \ENDWHILE
    \STATE $\vertex=A.\textrm{pop}()$
    \STATE $V'.\textrm{push}(\vertex)$
    \STATE Pop non-quantizable vertices/layers from $V'$
    \IF{|V'|>0}
        \STATE $V.\textrm{append}(V')$
    \ENDIF
\ENDWHILE
\RETURN $V$
\end{algorithmic}
\end{algorithm}

\section{Additional experimental results}\label{sec:additional_results}

\subsection{Per task: gained time based on measurements}

\begin{figure}[htbp]
    \centering
    \begin{subfigure}[b]{0.47\textwidth}
        \centering
        \includegraphics[width=0.7\textwidth]{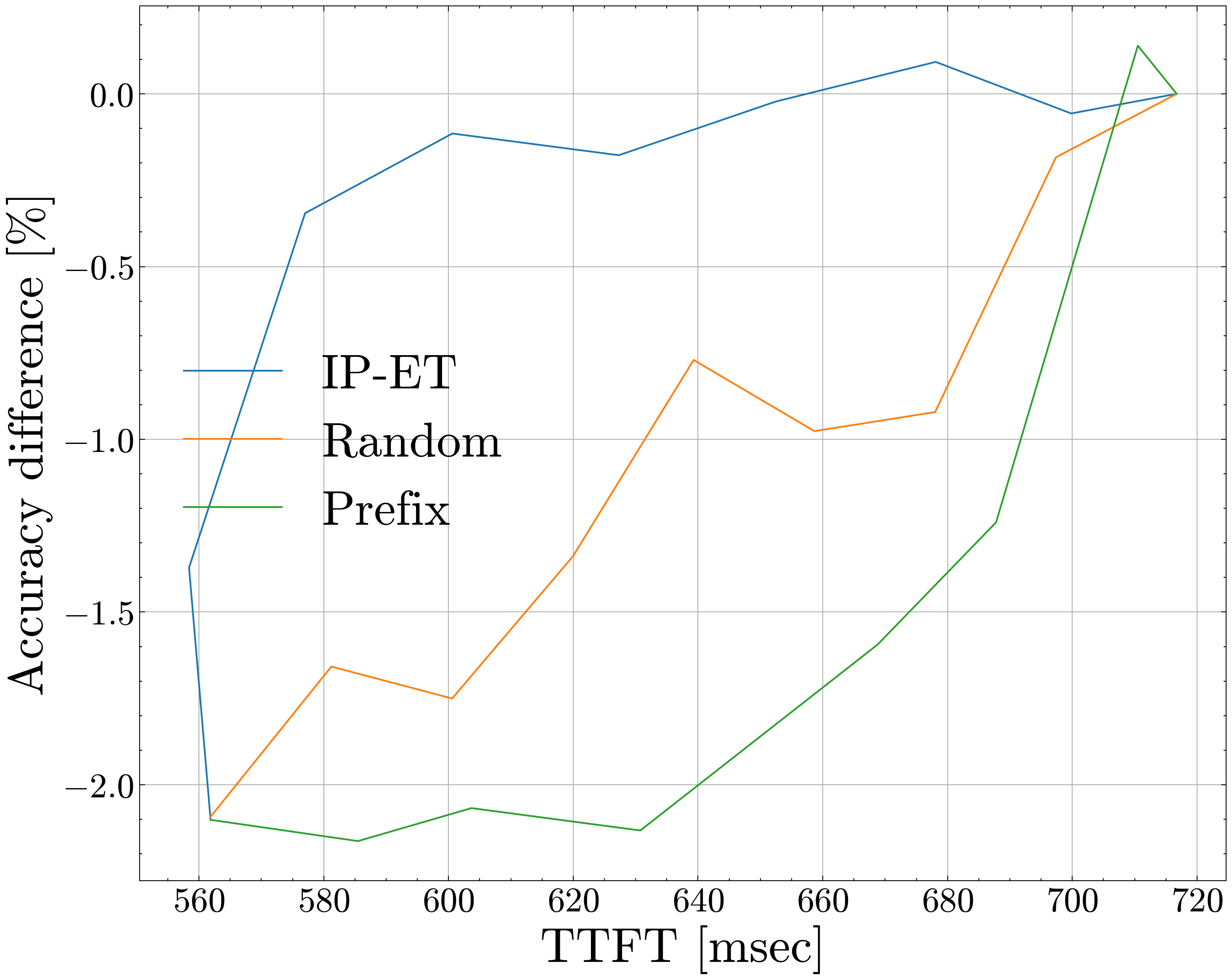}
        \caption{\ac{1B} - HellaSwag} \label{fig:1B/Measured/hellaswag_acc_norm_vs_ttft}
    \end{subfigure}
    \hfill
    \begin{subfigure}[b]{0.47\textwidth}
        \centering
        \includegraphics[width=0.7\textwidth]{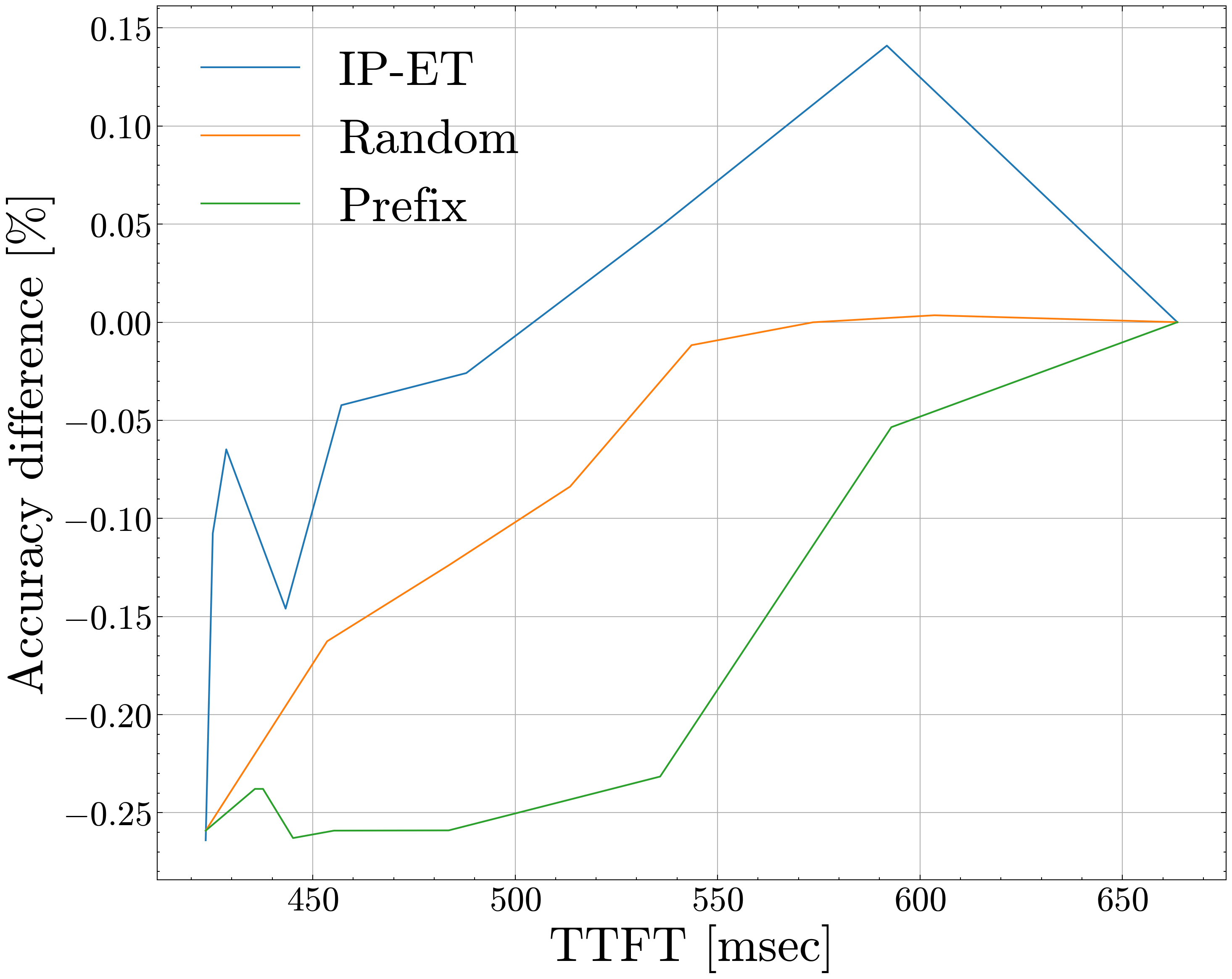}
        \caption{\ac{8B} - HellaSwag} \label{fig:8B/Measured/hellaswag_acc_norm_vs_ttft}
    \end{subfigure}
    
    \vspace{0.3cm}
    
    \begin{subfigure}[b]{0.47\textwidth}
        \centering
        \includegraphics[width=0.7\textwidth]{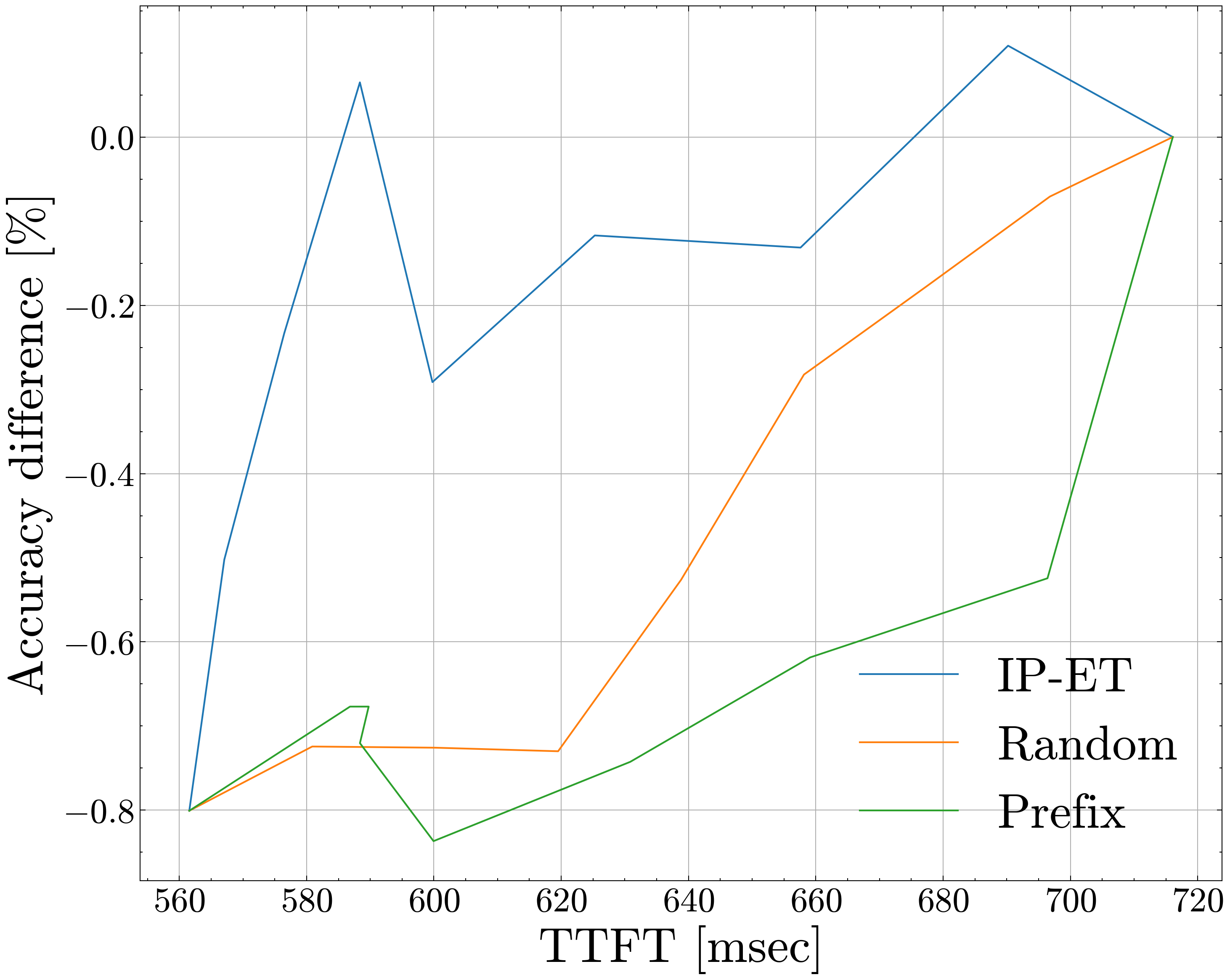}
        \caption{\ac{1B} - PIQA} \label{fig:1B/Measured/piqa_acc_norm_vs_ttft}
    \end{subfigure}
    \hfill
    \begin{subfigure}[b]{0.47\textwidth}
        \centering
        \includegraphics[width=0.7\textwidth]{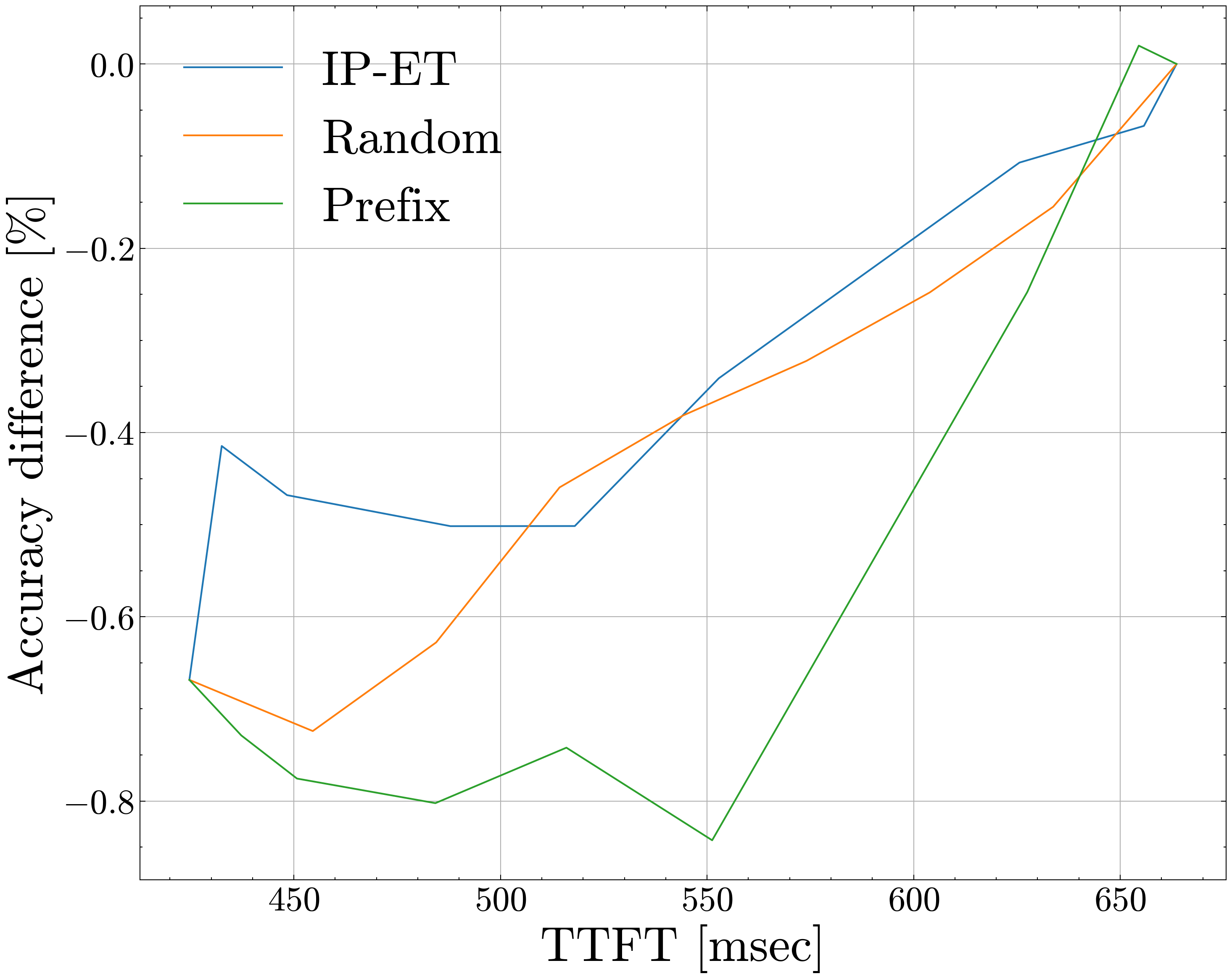}
        \caption{\ac{8B} - PIQA} \label{fig:8B/Measured/piqa_acc_norm_vs_ttft}
    \end{subfigure}

    \vspace{0.3cm}
    
    \begin{subfigure}[b]{0.47\textwidth}
        \centering
        \includegraphics[width=0.7\textwidth]{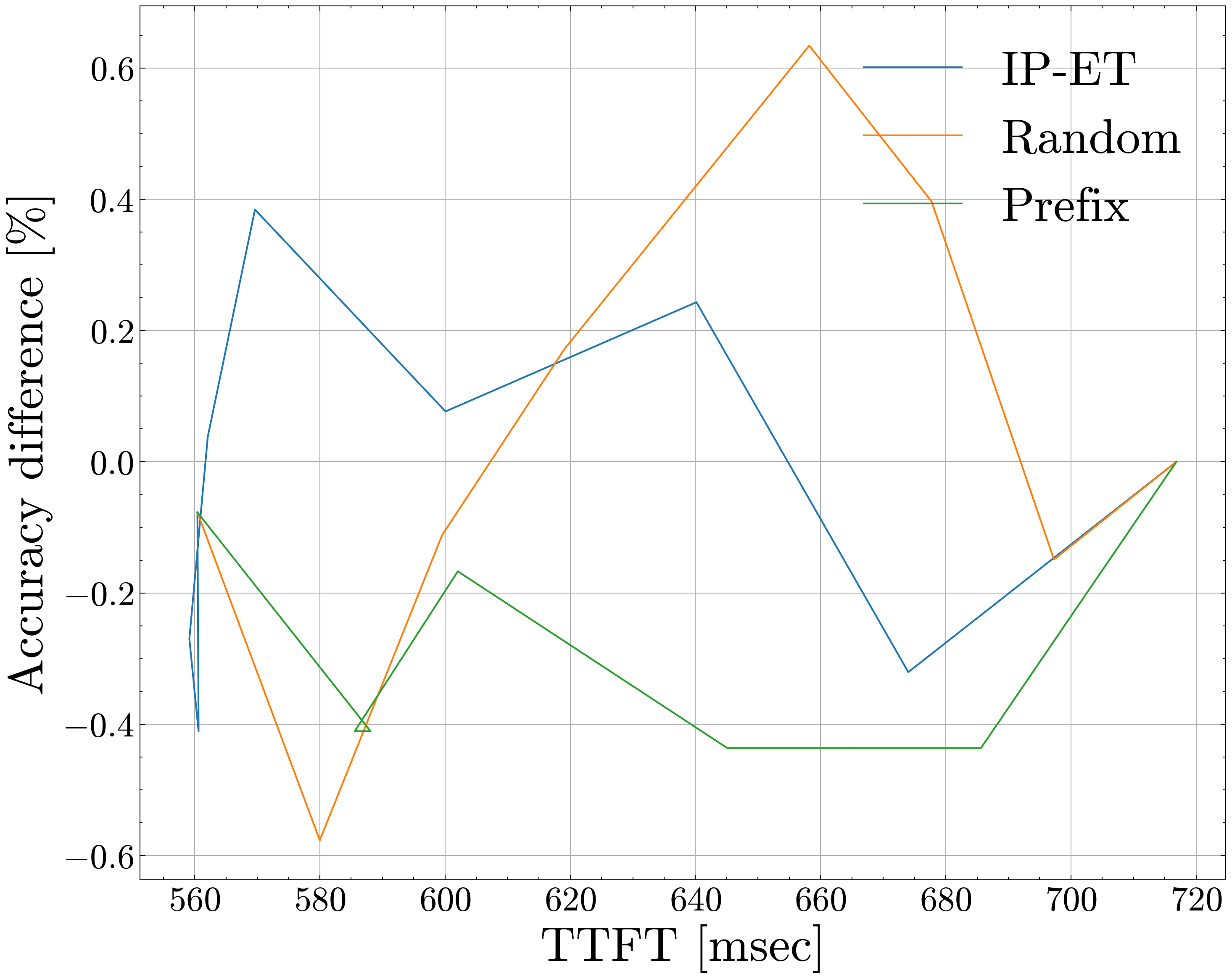}
        \caption{\ac{1B} - Winogrande} \label{fig:1B/Measured/winogrande_acc_vs_ttft}
    \end{subfigure}
    \hfill
    \begin{subfigure}[b]{0.47\textwidth}
        \centering
        \includegraphics[width=0.7\textwidth]{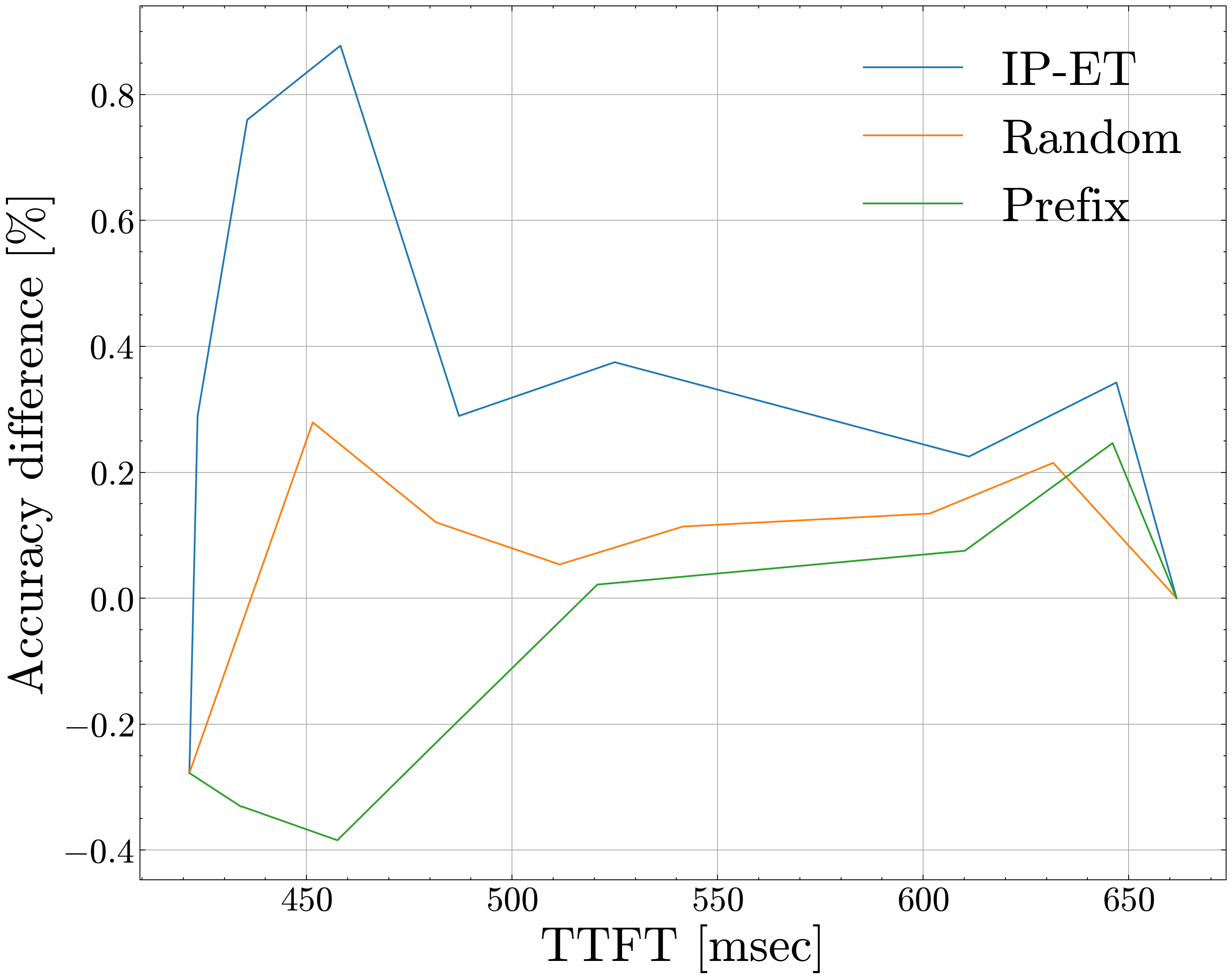}
        \caption{\ac{8B} - Winogrande} \label{fig:8B/Measured/winogrande_acc_vs_ttft}
    \end{subfigure}

    \vspace{0.3cm}
    
    \begin{subfigure}[b]{0.47\textwidth}
        \centering
        \includegraphics[width=0.7\textwidth]{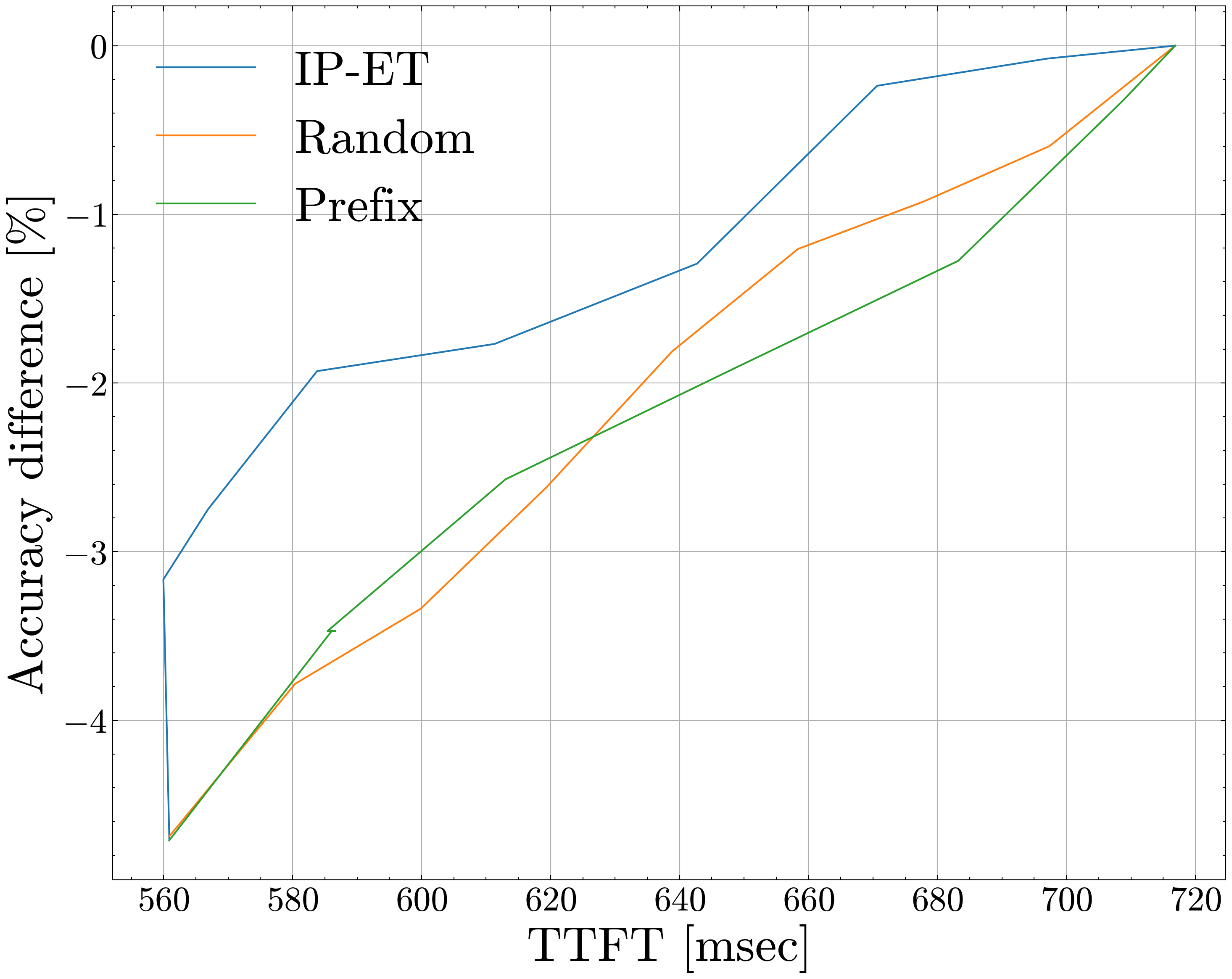}
        \caption{\ac{1B} - LAMBADA accuracy} \label{fig:1B/Measured/lambada_openai_acc_vs_ttft}
    \end{subfigure}
    \hfill
    \begin{subfigure}[b]{0.47\textwidth}
        \centering
        \includegraphics[width=0.7\textwidth]{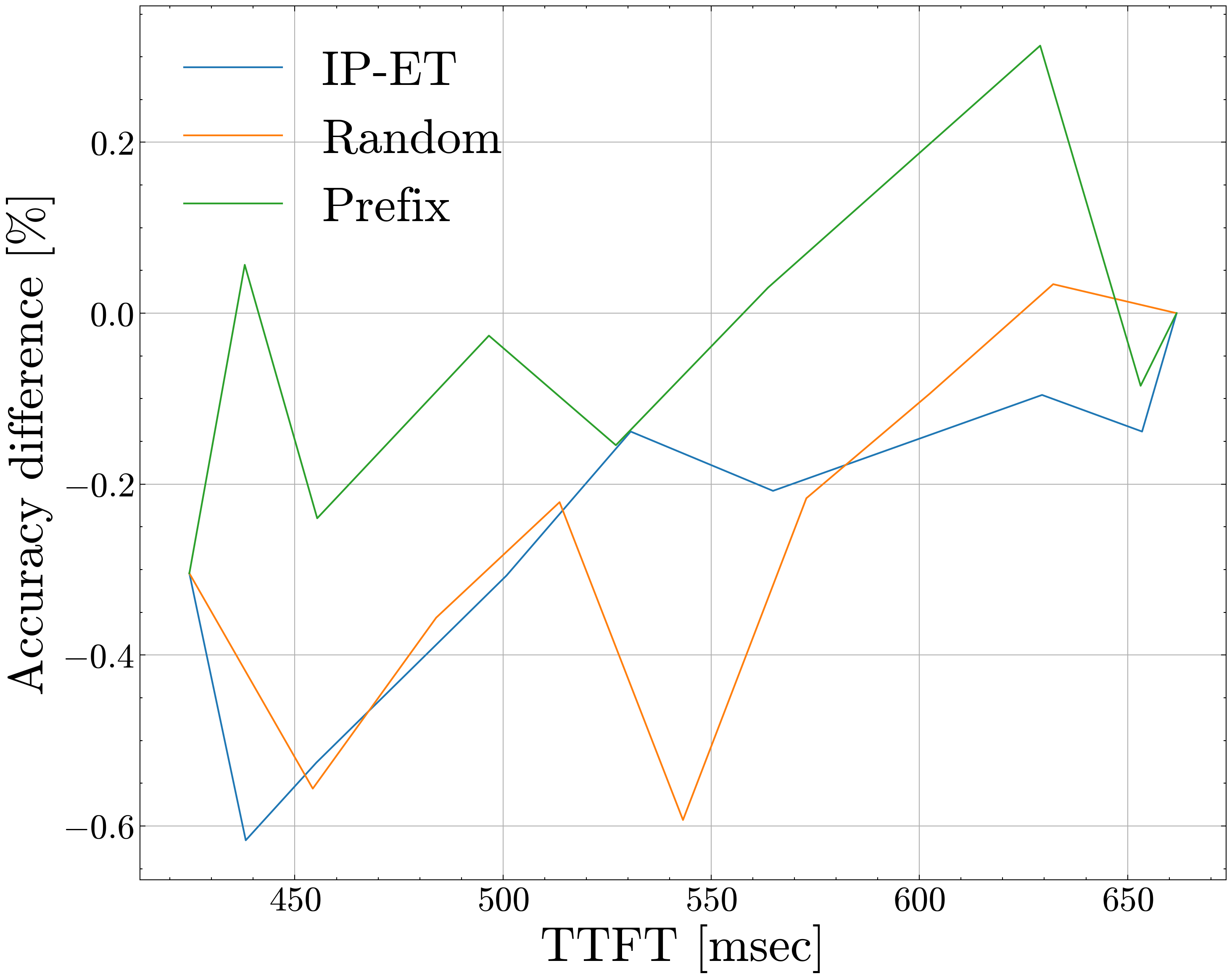}
        \caption{\ac{8B} - LAMBADA accuracy} \label{fig:8B/Measured/lambada_openai_acc_vs_ttft}
    \end{subfigure}
    
    \vspace{0.3cm}
    
    \begin{subfigure}[b]{0.47\textwidth}
        \centering
        \includegraphics[width=0.7\textwidth]{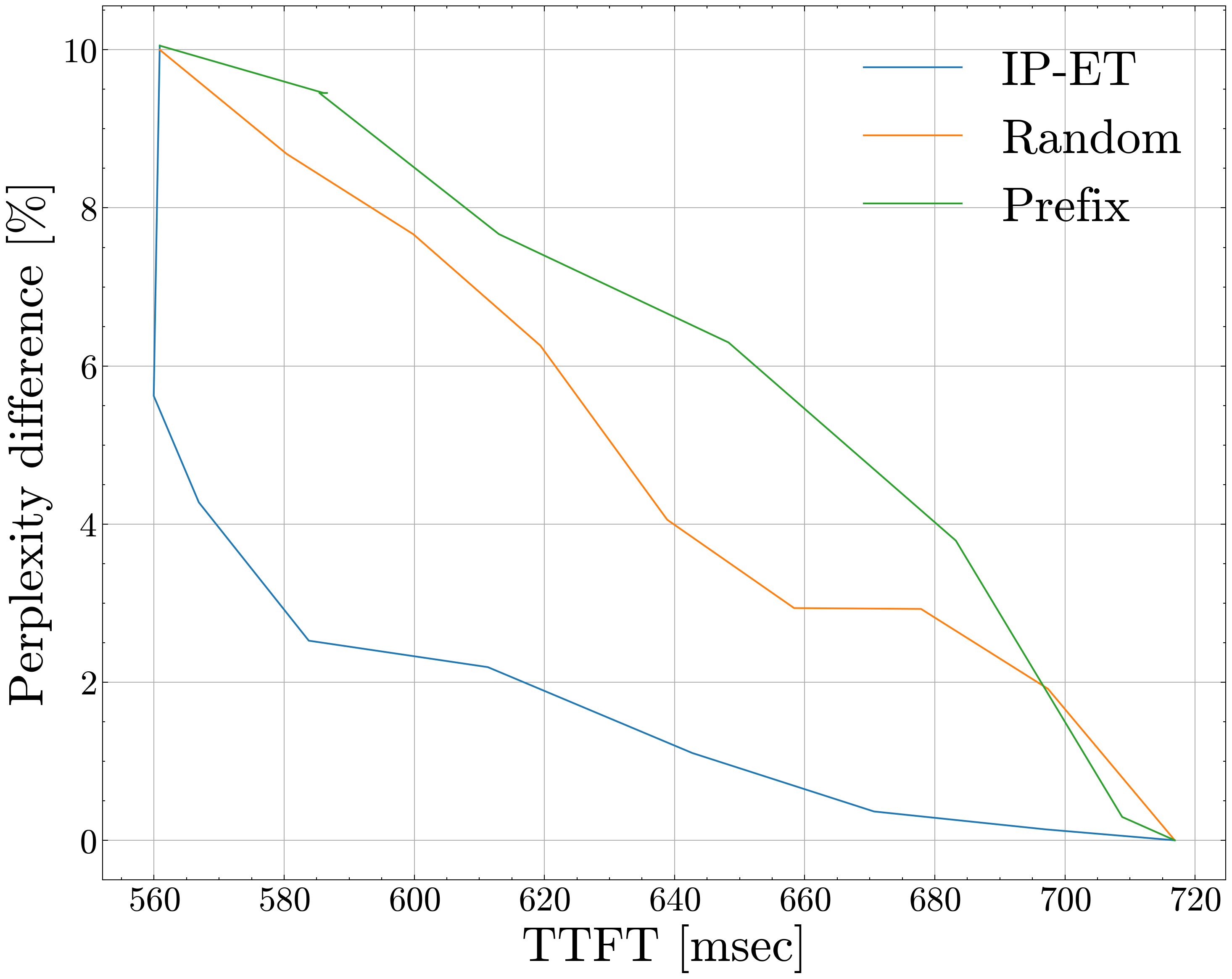}
        \caption{\ac{1B} - LAMBADA perplexity} \label{fig:1B/Measured/lambada_openai_ppl_vs_ttft}
    \end{subfigure}
    \hfill
    \begin{subfigure}[b]{0.47\textwidth}
        \centering
        \includegraphics[width=0.7\textwidth]{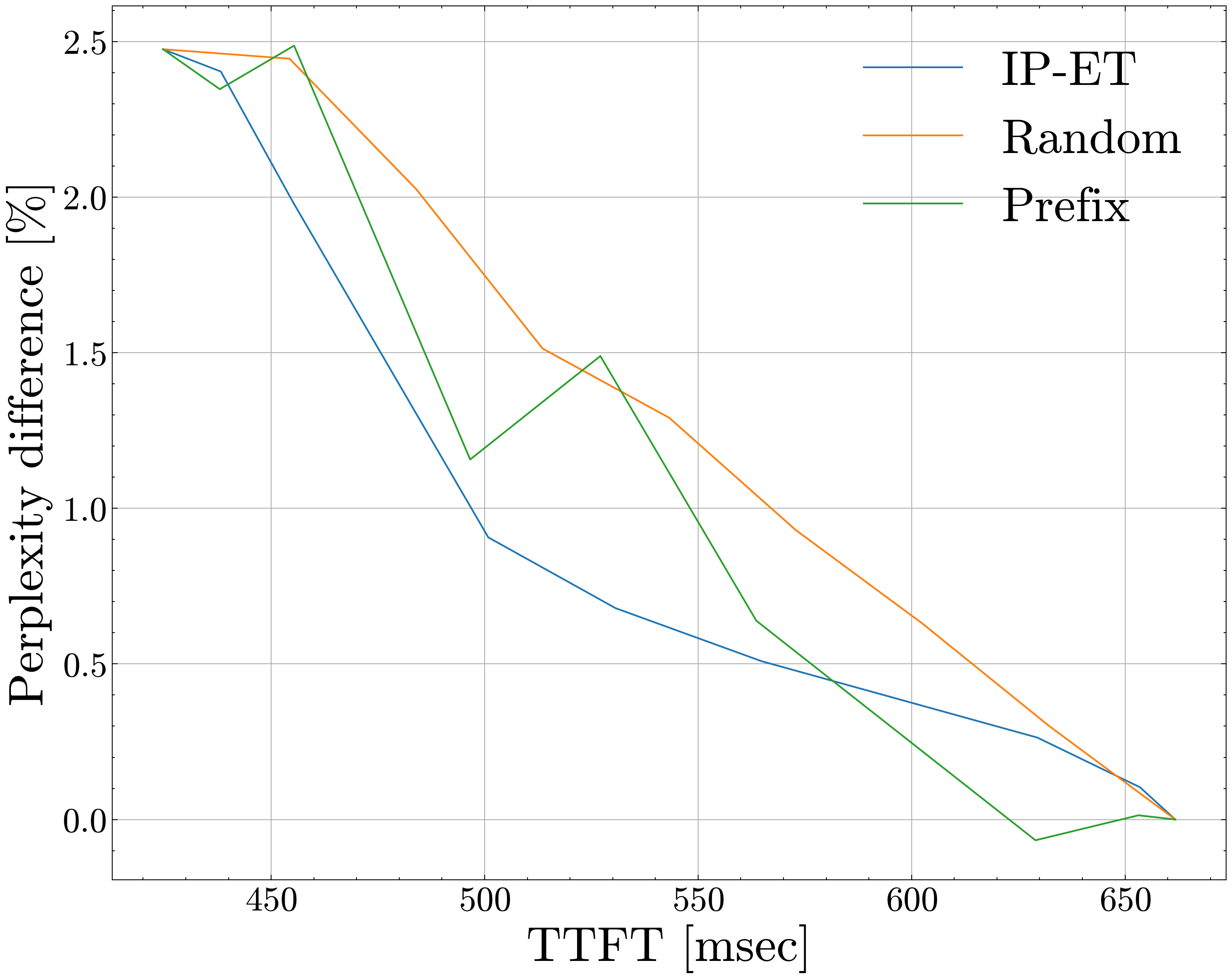}
        \caption{\ac{8B} - LAMBADA perplexity} \label{fig:8B/Measured/lambada_openai_ppl_vs_ttft}
    \end{subfigure}

    \caption{Per task accuracy/perplexity difference vs. TTFT. Comparing layer selection strategies for quantization (\ac{IP-ET}, Random, Prefix}
    \label{fig:per-task-measured}
\end{figure}

Figure \ref{fig:per-task-measured} reports for each individual task the accuracy difference (relatively to BF16) as a function of \ac{TTFT}. The proposed \ac{IP-ET} outperforms Random and Prefix strategies on most of the settings, particularly in the \ac{1B} model. For example, in
HellaSwag using the \ac{1B} model (Figures \ref{fig:1B/Measured/hellaswag_acc_norm_vs_ttft} and \ref{fig:8B/Measured/hellaswag_acc_norm_vs_ttft}), \ac{IP-ET} shows a significant advantage across all \ac{MP} configurations. 

In LAMBADA using the \ac{8B} model (Figure \ref{fig:1B/Measured/lambada_openai_acc_vs_ttft}), Prefix yields higher accuracy, but \ac{IP-ET} achieves lower perplexity (Figure \ref{fig:1B/Measured/lambada_openai_ppl_vs_ttft}), highlighting that loss-based optimization (which is correlated to perplexity) doesn’t necessarily translate to accuracy gains. 

In Winogrande on the \ac{1B} model (Figure~\ref{fig:1B/Measured/winogrande_acc_vs_ttft}) is particularly noisy, as reflected by large standard deviations in Table~\ref{tab:merged_configs_avg} which can explain the reason \ac{IP-ET} shows no clear advantage. However, the rapid rise of the blue line indicates that \ac{IP-ET}, achieves good accuracy by not quantizing only a few layers.

\subsection{Gained time based on number of \acp{MAC}}

\begin{figure}[H]
    \centering
    \begin{subfigure}{0.49\textwidth}
        \centering
        \includegraphics[width=\linewidth]{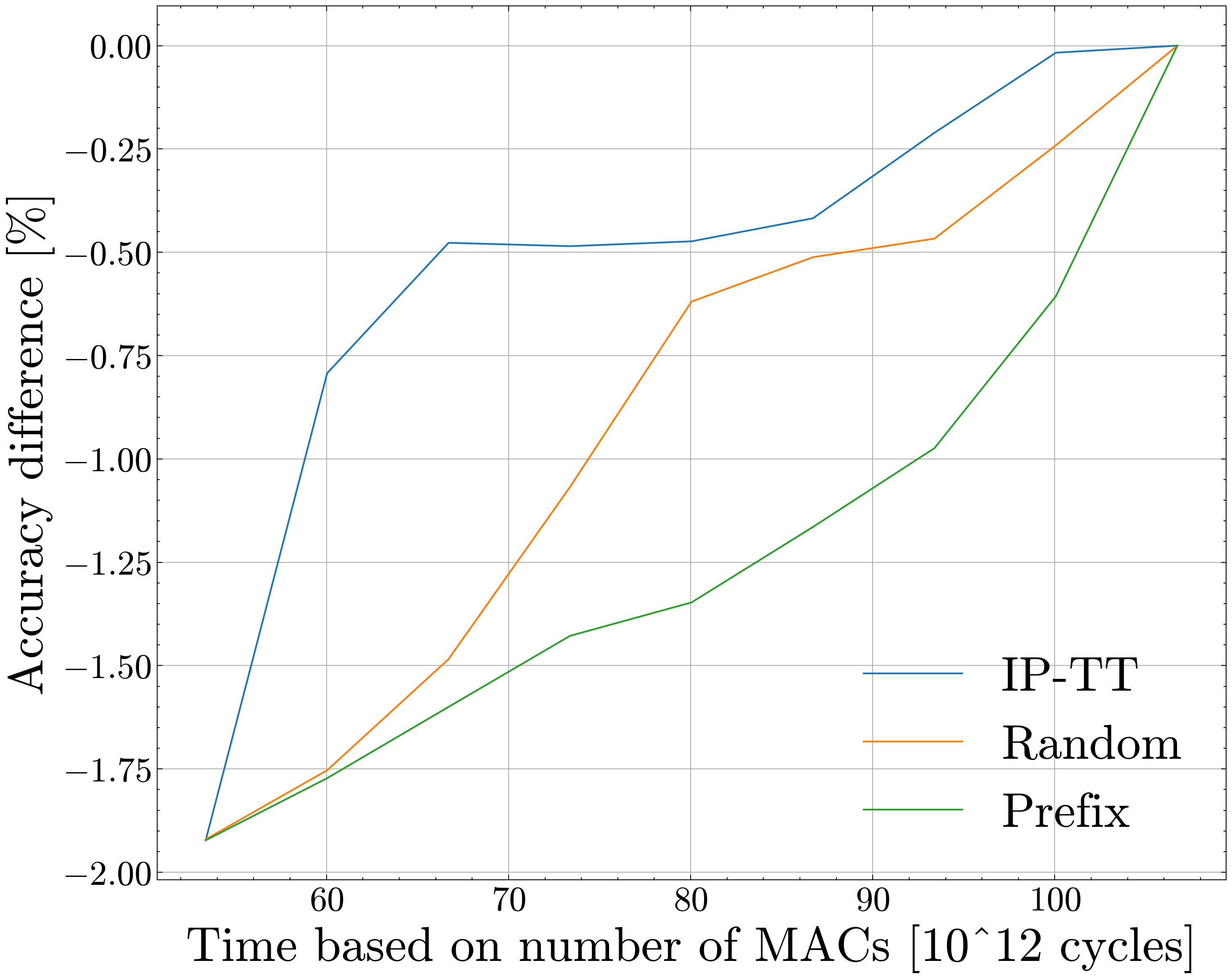}
        \caption{\ac{1B}}
        \label{fig:1B/MACs/acc_avg_vs_macs}
    \end{subfigure}
    \hfill
    \begin{subfigure}{0.49\textwidth}
        \centering
        \includegraphics[width=\linewidth]{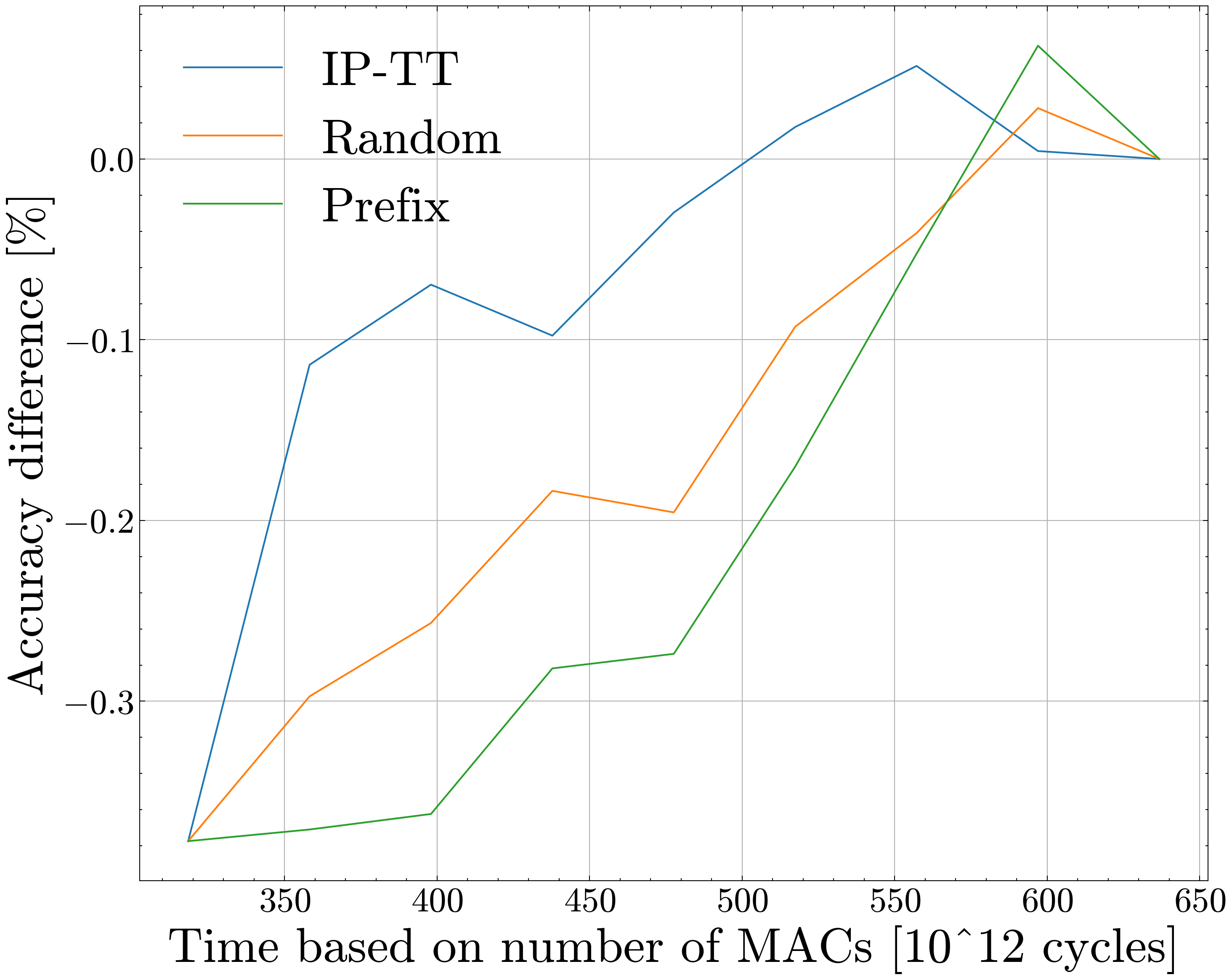}
        \caption{\ac{8B}}
        \label{fig:8B/MACs/acc_avg_vs_macs}
    \end{subfigure}
    \caption{Average accuracy difference [\%] vs. time based on number of MACs [cycles], across HellaSwag, LAMBADA, Winogrande, and PIQA. Comparing layer selection strategies for quantization (\ac{IP-TT}, Random, Prefix).}
    \label{fig:macs-comparison}
\end{figure}

Figure \ref{fig:macs-comparison} shows the tradeoff between accuracy and theoretical compute time, measured by MACs. The x-axis denotes the theoretical time gain based on the number of MAC operations as defined in Sec.~\ref{subsec:theoretical_time}. While the y-axis reports accuracy difference relative to BF16, averaged across tasks. Our \ac{IP-TT} (blue) consistently outperforms Random (orange) and Prefix (green) strategies, achieving a smaller accuracy degradation on both model sizes.

\subsection{Gained memory}

\begin{figure}[H]
    \centering
    \begin{subfigure}{0.49\textwidth}
        \centering
        \includegraphics[width=\linewidth]{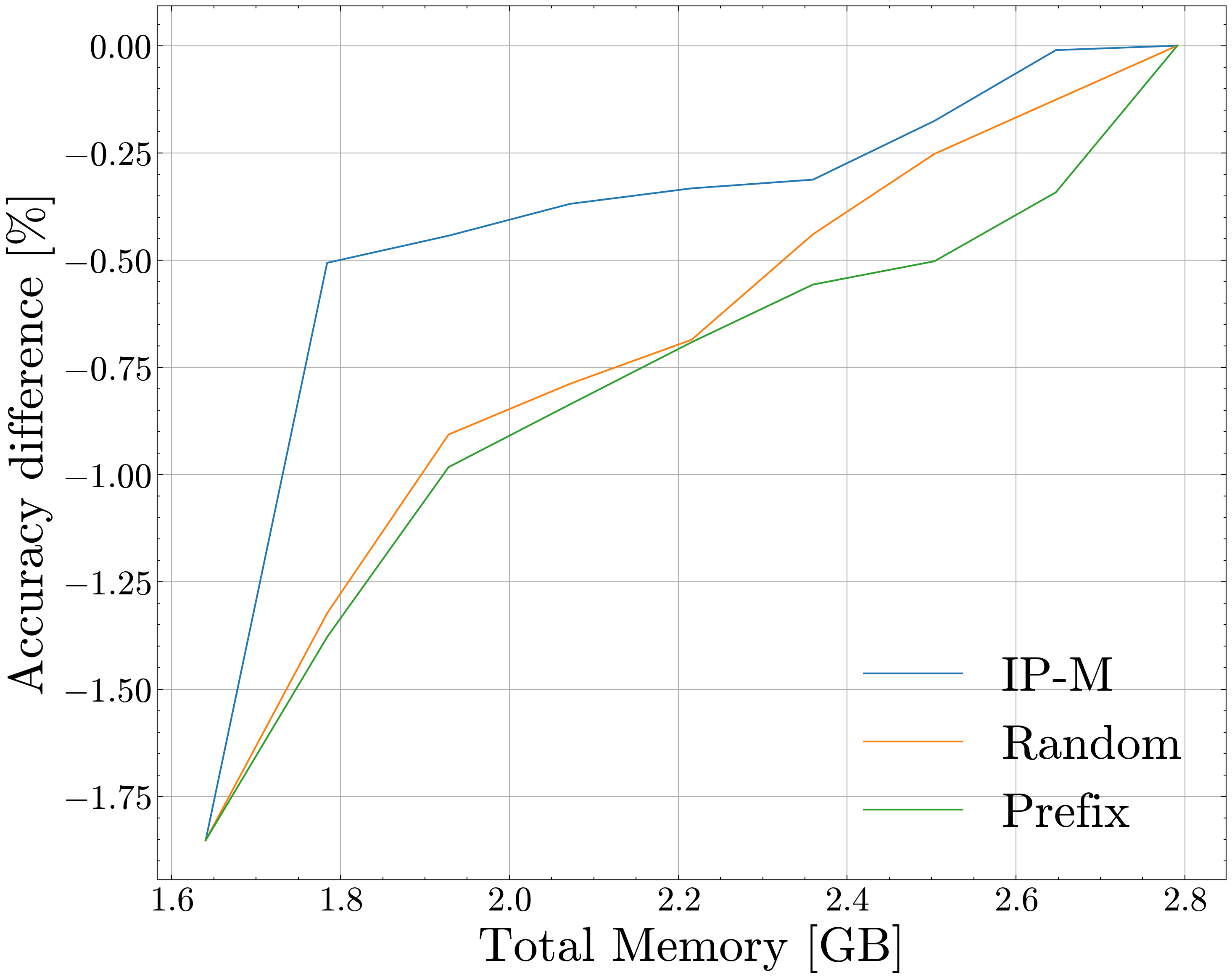}
        \caption{\ac{1B}}
        \label{fig:1B/Memory/acc_avg_vs_total_memory}
    \end{subfigure}
    \hfill
    \begin{subfigure}{0.49\textwidth}
        \centering
        \includegraphics[width=\linewidth]{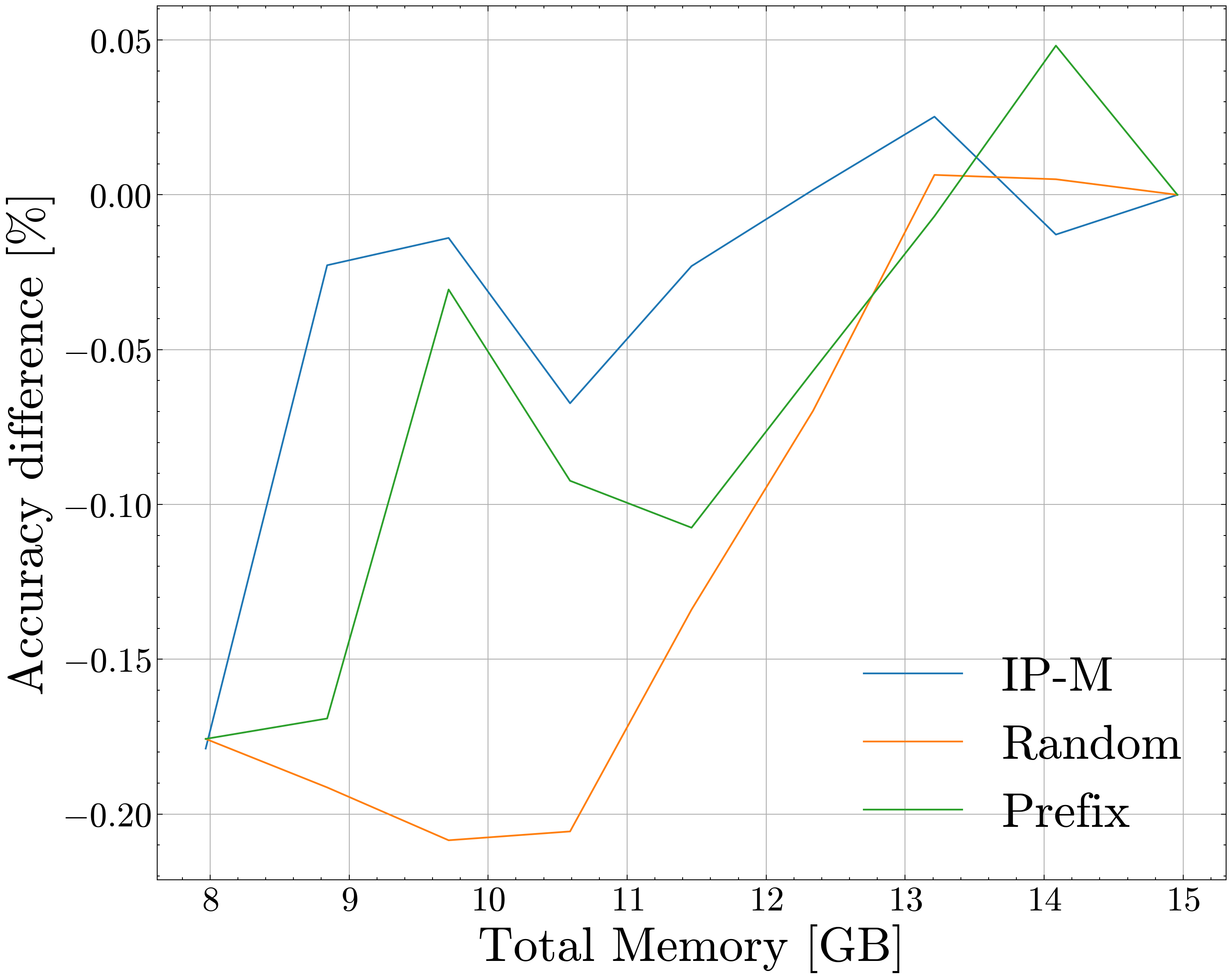}
        \caption{\ac{8B}}
        \label{fig:8B/Memory/acc_avg_vs_total_memory}
    \end{subfigure}
    \caption{Average accuracy difference [\%] vs. total memory across HellaSwag, LAMBADA, Winogrande, and PIQA. Comparing layer selection strategies for quantization (\ac{IP-M}, Random, Prefix).}
    \label{fig:memory-comparison}
\end{figure}

Figure \ref{fig:memory-comparison} shows the tradeoff between accuracy and total model's memory. The x-axis values were calculated by subtracting BF16 model's total memory and the memory gain (defined in Sec.~\ref{subsec:theoretical_time}) of each configuration. While the y-axis reports accuracy difference relative to BF16, averaged across tasks. 

For the \ac{1B} model (Figure \ref{fig:1B/Memory/acc_avg_vs_total_memory}), \ac{IP-TT} (blue) consistently outperforms Random (orange) and Prefix (green) strategies, achieving lower accuracy loss for a given memory budget. 

For the \ac{8B} model (Figure \ref{fig:8B/Memory/acc_avg_vs_total_memory}), \ac{IP-TT} also performs better than other strategies, though the margin is small; notably, the initial FP8 configuration results in less than 0.2\% accuracy difference range, since only linear layers are quantized in these experiments. All \ac{8B}'s \ac{MP} configurations yield averaged accuracy difference close to zero.

\end{document}